\documentclass[acmsmall]{acmart}
\AtBeginDocument{%
  \providecommand\BibTeX{{%
    \normalfont B\kern-0.5em{\scshape i\kern-0.25em b}\kern-0.8em\TeX}}}

\setcopyright{acmcopyright}
\copyrightyear{2021}
\acmYear{2021}
\acmDOI{}

\acmJournal{JACM}
\acmVolume{X}
\acmNumber{X}
\acmArticle{X}
\acmMonth{12}

\usepackage{multirow}

\usepackage{cleveref}

\crefformat{section}{\S#2#1#3}
\crefformat{subsection}{\S#2#1#3}
\crefformat{subsubsection}{\S#2#1#3}
\crefrangeformat{section}{\S\S#3#1#4 to~#5#2#6}
\crefmultiformat{section}{\S\S#2#1#3}{ and~#2#1#3}{, #2#1#3}{ and~#2#1#3}

\begin{document}

\title{Automatic Related Work Generation: A Meta Study}

\author{Xiangci Li}
\email{lixiangci8@gmail.com}
\author{Jessica Ouyang}
\email{jessica.ouyang@utdallas.edu}
\affiliation{%
  \institution{University of Texas at Dallas}
  \streetaddress{800 W Champbell Rd.}
  \city{Richardson}
  \state{Texas}
  \country{USA}
  \postcode{75080}
}

\renewcommand{\shortauthors}{Li and Ouyang}

\begin{abstract}

Academic research is an exploration activity to solve problems that have never been resolved before. By this nature, each academic research work is required to perform a literature review to distinguish its novelties that have not been addressed by prior works. In natural language processing, this literature review is usually conducted under the “Related Work” section. The task of automatic related work generation aims to automatically generate the “Related Work” section given the rest of the research paper and a list of 
cited papers. Although this task was proposed over 10 years ago, it received little attention until very recently, when it was cast as a variant of the scientific multi-document summarization problem. However, even till today, the problems of automatic related work and citation text generation are not yet standardized. In this survey, we conduct a meta-study to compare the existing literature on related work generation from the perspectives of problem formulation, dataset collection, methodological approach, performance evaluation, and future prospects to provide the reader insight into the progress of the state-of-the-art studies, as well as and how future studies can be conducted. We also survey relevant fields of study that we suggest future work to consider integrating.

\end{abstract}



\maketitle

\section{Introduction}

Academic research is an exploration activity to solve problems that have never been resolved before. By this nature, each academic research work must sit at the frontier of the field, and present novelties that have not been addressed by prior works. While the format may vary among different fields, in order to convince the reviewers and readers of the novelty of the current work, the authors must perform a literature review to compare their work with the prior works. In natural language processing, this literature review is usually conducted under the “Related Work” section. Since each author is obligated to review the relevant prior work in their field, many of which are shared among papers with the same task or topic, many related work sections in the same field are similar in both content and format. Moreover, the related work section is usually a stand-alone section that is relatively independent of the main thread of the paper and is usually written after the authors finish the main part of the paper. Therefore, it is a natural motivation to develop a system for generating related work sections automatically.

As \citet{hoang-kan-2010-towards} envision, the automatic related work generation task assists authors in creating a draft of a related work section. To generated the topic-biased related work, the task is a variant of multi-document summarization problems that takes a target scientific document for which a related work section needs to be drafted, as well as a list of papers to be cited as the input. As \citet{agarwal2011scisumm} point out, multi-document summarization of scientific articles has unique characteristics compared to summarization of generic texts: summarizing multiple scientific articles requires understanding different arguments for each study, even if their research directions are the same. Moreover, an accurate description of a literature has to consider and combine all other different viewpoints toward the literature.
The complexity of the concepts and the use of rare, technical terms in scientific articles further makes this task challenging \cite{yasunaga2019scisummnet}. 

However, automatic related work generation is not only about the multi-document summarization in the scientific domain. The challenge of automatic related work generation depends on the development of related, natural language understanding subtasks, such as citation and discourse analysis. Since related work generation not only requires the understanding of the main idea of a single document but the relationships and interactions among multiple documents, only when such relationships are modeled can we accurately generate a proper summary. 
Therefore, it is crucial to have a higher-level view that includes tasks and techniques that are beyond the central summarization task. Unfortunately, the existing studies have not reached this step yet.

Although the problem of automatic related work generation was proposed over 10 years ago, this problem received little attention until very recently, when a few works revisited this problem with neural network-based systems \cite{abura2020automatic, xing2020automatic, ge-etal-2021-baco, chen-etal-2021-capturing, luu-etal-2021-explaining, deng2021automatic}. There are only a few existing papers that explicitly study related work generation. Further, the problems of automatic related work and citation text generation are not yet standardized, as each work has slightly different assumptions about the task, datasets used, and methods of evaluation, so that it is hard to compare the existing work in parallel. In this survey, we present a critical review of the state-of-the-art studies of the task of automatic related work generation by comparing from multiple perspectives such as problem formulation, dataset collection, methodological approaches, experimental results and evaluation across the limited number of available prior works. 
We also survey relevant fields of study such as citation analysis to inspire readers to conduct future studies on related work generation by considering and incorporating these related tasks. 

We organize the rest of this paper as a meta-study of the task of related work generation, by providing an overview of the directly related works (\cref{sec:overview}), and a series of parallel comparisons of these works from the perspective of problem formulation (\cref{sec:problem_formulation}), dataset collection (\cref{sec:dataset}), methodological approaches (\cref{sec:approaches}), experimental results and evaluation (\cref{sec:evaluations}), proposed future work (\cref{sec:future_work}), then we review related tasks that potentially benefit related work generation (\cref{sec:related_studies}), and finally we perform a general discussion (\cref{sec:discussion}).

\section{Terminologies}
\paragraph{Related Work Section.}
A stand-alone literature review section in research papers, which is typically seen in natural language processing (NLP) papers. In other fields, such as information studies, it may be referred to as simply a ``literature review.''

\paragraph{Related Work Generation.}
The task of automatically generating related work sections as related work generation. Since this process largely involves summarization, it was often referred to as ``related work summarization'' in early literature, such as \citet{hoang-kan-2010-towards}. 

\paragraph{Target Paper.}
The paper whose related work section is to be generated. It is also known as the citing paper for citation analysis.

\paragraph{Cited Paper.}
A paper cited by the target paper. It is also referred as a ``reference paper'' in some literature.

\section{Overview of Previous Related Work Generation} \label{sec:overview}
There is only a limited number of previous studies about related work generation, in this section, we provide an overview of the studies before analyzing them from different aspects in the following sections.
We introduce the extractive approaches first, then abstractive approaches, which is mostly in the chronological order.
 
\subsection{Extractive Related Work Generation}

\subsubsection{First Work of Related Work Generation} \hfill \break
\label{sec:hoang2010}
The problem of automatic related work generation is first proposed by \citet{hoang2010towards}. They envision a related work summarization task that assists in creating a draft of the related work section. They frame it as a topic-biased, multi-document summarization problem that takes a target scientific document for which a related work section needs to be drafted and a list of cited papers as the inputs, and the output goal is to create a related work section that contextually describes the related prior works in relationship to the scientific document at hand. They identify the main steps of related work generation: (1) Finding relevant documents; (2) Identifying the salient aspects of these documents in relation to the current work worth summarizing; (3) Generating the topic-biased summary. Their work focus on the last part that creates a related work section given a structured input of the topics for a summary.

\citet{hoang2010towards} start from an existing topic hierarchy tree of the target related work as the input. Each node of the tree represents a topic such as ``text classification'', and sub-topic nodes are derived via tree edges from their parent nodes. \citet{hoang2010towards} propose a largely heuristic system called ReWoS that takes different strategies for general background information and specific author contribution summarizations. For the general content, they generate ``indicative'' sentences with templates, and use Term Frequency $\times$ Inverse Sentence Frequency (TF $\times$ ISF) scores \cite{otterbacher2005using} to extract sentences from each cited article as ``informative'' sentences. On the other hand, specific author contributions are also generated in an extractive manner with additional context modeling and weighting rules. Finally, the pronouns such as ``we'' in the extracted sentences are replaced with the corresponding citation references.

As the first system proposed for related work generation task, \citet{hoang2010towards}'s RoWoS obtains more than 3.5 out of 5 points in correctness, novelty, and usefulness scores evaluated by humans, which seems to be even better than recent neural network-based systems. However, they only study 20 articles which is the smallest scale ever. Moreover, their system heavily relies on the pre-built topic tree, while the construction of the topic tree is also a challenging task. 

\subsubsection{Extractive Related Work Generation with Topic Modeling} \hfill \break
\citet{hu2014automatic} inherit the task definition of related work generation from \citet{hoang2010towards} and they are the first work that propose a system for the full task of related work generation. \citet{hu2014automatic} take the abstract and introduction sections of the target paper and the abstract, introduction, related work, and conclusion sections of the reference papers as the inputs to their system. They leverage a Probabilistic Latent Semantic Analysis (PLSA) model \cite{hofmann1999probabilistic}, a kind of topic model, to group sentence sets of the target paper and the cited papers into topic-biased clusters. Then they train two feature-based Support Vector Regression (SVR) models for sentences in the target paper and sentences in reference papers separately to learn the sentence importance scores to distinguish sentences that describe the author's own work and the sentences that describe the related works. Finally, they formulate the generation problem as an optimization problem by searching for a set of sentences to optimize the objective function to generate the related work section in an extractive fashion. 

\citet{chen2019automatic} further inherit the setting of \citet{hu2014automatic} to generate the related work section given the title, abstract, introduction, and conclusion sections of the target paper and a list of cited papers, as well as the papers that cite the cited papers. The keywords in the title of the paper being written are used as the topics for the related work section. They construct a minimum Steiner tree of the keywords and generate a related work section by extracting the sentences from the papers that co-cite the same cited papers with the target paper to cover the minimum Steiner tree. They use a discriminative feature graph for the sentence selection. Since the set cover problem is an NP-hard problem \cite{fraer1999minimum}, they propose a greedy approximation algorithm by choosing the sentence that contains the largest number of uncovered elements each time step. The ROUGE score-based \cite{lin2004rouge} automatic evaluation show that \citet{chen2019automatic}'s RWS-Cit system outperforms \citet{hoang-kan-2010-towards}'s RoWoS that \citet{chen2019automatic} replicate.

\subsubsection{Identifying Cited Text Span for Extractive Related Work Generation}
\hfill \break
In addition to leveraging a topic model, \citet{wang2019toc} also consider explicitly extracting Cited Text Spans (CTS), which are the matched text spans in the cited paper that are most related to a given citation. Their topic model, QueryTopicSum extends TopicSum \cite{haghighi2009exploring}, which is designed for general domain multi-document summarization, by capturing the relationship between the target paper and the cited papers. In parallel to topic modeling, they also consider citation sentences from the papers that co-cite the cited papers in the target related work section, in addition to the common target paper context and the cited papers. They use a two-layered ensemble model to extract CTS as a classification task. Finally, a greedy algorithm selects the candidate sentences to generate the related work section.

\subsubsection{Sentence Extraction and Reordering}
\citet{deng2021automatic}'s SERGE system is the only extractive related work generation approach based on the neural network. They first apply a ROUGE-based retrieval approach \cite{yasunaga2019scisummnet} to automatically create a training set by retrieving salient sentences from the abstract and conclusion sections of the cited papers in ScisummNet \cite{yasunaga2019scisummnet} to maximize the ROUGE score \cite{lin2004rouge} with citation sentences in the target papers. Then they train an ensemble model with BERT \cite{devlin2018bert} and keyword-based bag-of-word model for the salient sentence extraction. Finally, they simplify the sentence reordering problem to the next sentence prediction problem, implemented by a fine-tuned MobileBERT \cite{sun2020mobilebert}.

Because their SERGE system does not directly require citation context during the training and inference steps, they claim that their approach does not suffer from the delay of citing new publications.

\subsection{Abstractive Related Work Generation}
Due to the development of neural network-based models \cite{see-etal-2017-get, lewis2019bart, 2020t5} for abstractive summarization tasks, recent work of related work generation typically takes a bottom-up strategy to generate fine-grained elements such as tokens and sentences first, and use them as building blocks for the complete related work section in the future studies. These works may not mention ``related work generation'' as their task, but instead phrase them as smaller tasks such as ``citation text generation'' \cite{xing2020automatic, ge-etal-2021-baco} or ``explaining relationships between scientific documents'' \cite{luu-etal-2021-explaining}, but we consider all these works as variations for related work generation.

\subsubsection{Neural Network-Based Abstractive Related Work Generation} \hfill \break
As a pioneering work of applying abstractive text generation to the related work generation task, \citet{abura2020automatic} apply the Pointer-Generator Network \cite{see-etal-2017-get} and Transformers \cite{vaswani2017attention} to generate citation sentences given the concatenation of the cited paper's title and abstract. Although they do not propose any new model architecture, their study introduces the abstractive summarization approach to the field of related work generation.

\subsubsection{Abstractive Citation Text Generation Given Contexts} \hfill \break
In parallel to \citet{abura2020automatic}, \citet{xing2020automatic} is the first work that explicitly considers citation contexts for abstractive citation text generation. They collect a citation text generation dataset based on the ACL Anthology Network corpus (AAN) \cite{radev2013acl}. They train a BERT \cite{devlin2018bert} model on a small manually annotated dataset to label implicit citation sentences from each citation text block. Then they extend the Pointer-Generator Network \cite{see-etal-2017-get} to propose the PTGEN-Cross model with a cross-attention mechanism that takes both the citation context and the cited paper's abstract as the input. Their generation target is the citation sentence with a single cited document.

\citet{xing2020automatic} take several effective approaches such as the combination of manual and automated explicit and implicit citation sentence labeling, and neural network-based abstractive citation text generation. However, they only consider sentences with a single citation, and their proposed model, PTGEN-Cross also hard-codes the setting of two input sources, instead of a variable number of input sources.

\subsubsection{Using External Information to Improve Citation Sentence Generation} \hfill \break
\citet{ge-etal-2021-baco} extend \citet{xing2020automatic}'s method by encoding citation network information using a graph attention network (GAT) \cite{Velickovic2018GraphAN}, in addition to encoding the citation context and the cited abstracts using a hierarchical bidirectional LSTM \cite{hochreiter1997long}. In addition to the citation sentence, their model also predicts the citation function and the salient sentences in the cited paper abstracts. All these three prediction tasks are trained jointly via multi-task learning \cite{caruana1997multitask}.

\citet{ge-etal-2021-baco}'s experiments demonstrate the effectiveness of additional background knowledge (citation network), salience estimation, and citation function to the performance of citation sentence generation. This also suggests the potential of improving the performance by introducing other additional learning signals. However, due to their inheritance of \citet{xing2020automatic}'s task setting, their approach shares the same limitations as \citet{xing2020automatic} that their system is only applicable to citation texts with a single citation.

\subsubsection{Capturing Relations Between Scientific Papers} \hfill \break
In a parallel line, \citet{chen-etal-2021-capturing} work on the problem of generating related work section paragraphs with two or more citations given multiple cited papers' abstracts. They derive their related work generation dataset from S2ORC \cite{lo-wang-2020-s2orc} and Delve \cite{akujuobi2017delve}, both of which are large datasets with a large connected citation graph. They also use Amazon Mechanical Turk to verify that the related work can be partly generated based on the given abstracts of the cited papers. 

Their main contribution is proposing a custom relation-aware related work generator (RRG). The RRG model has three main parts: (1) The hierarchical encoder implemented by a Transformer \cite{vaswani2017attention} encodes multiple input cited abstracts. They implement two types of positional encoding to encode document index and word index separately. (2) The relationship modeling part fuses encoded cited abstract representations to relation-aware document encoding and a polished relation graph. They customize a stacked Transformer-block-like \cite{vaswani2017attention} architecture, and each block contains a relation graph updater that consists of a multi-head attention mechanism \cite{vaswani2017attention} and an LSTM-like \cite{hochreiter1997long} gate. (3) Finally a Pointer-Generator Network-like \cite{see-etal-2017-get} module is used for generating related work.

\citet{chen-etal-2021-capturing} is the first work on generating related work paragraphs with multiple citations given multiple cited papers' abstracts. Their task setting is closer to the real related work section generation scenario.

\subsubsection{Explaining Relationships Between Scientific Documents} \hfill \break
In parallel to \citet{ge-etal-2021-baco} and \citet{chen-etal-2021-capturing}: \citet{luu-etal-2021-explaining} frame their task as using citation sentence as partial supervision for explaining relationships between two scientific documents.
They use citation sentences in the S2ORC dataset \cite{lo-wang-2020-s2orc} with a single reference that links back to the corpus as their dataset. They experiment with two approaches. The first approach is the neural network-based abstractive generation, by fine-tuning GPT2 \cite{radford2019language} on texts in the scientific domain (SciGPT2) and then further fine-tuned on the task dataset (S2ORC \cite{lo-wang-2020-s2orc}). The second approach is based on information retrieval (IR), which retrieves the citation sentence from the citing-cited paper pair closest to the target citing-cited paper pair. They further try information extraction techniques to extract named entities or terms from the cited papers to enrich the paper representation. To combat hallucinations and promote factual accuracy, they rank the candidate answers based on the entity matching between the candidate sentences and the cited document. 

Compared to the systems proposed above, \citet{luu-etal-2021-explaining} propose a simpler approach by further fine-tuning an existing model, which is easier to apply to future scenarios. However, the GPT2 \cite{radford2019language} base model only has a limited context window of 512 (base) or 1024 (large), which is a severe bottleneck for modeling the relationship between two documents. Their problem formulation of explaining the relationships between two documents is ingenious, but this setting does not apply to the real scenario where multiple documents may be cited in a single sentence.

\hfill \break

\noindent Next, we compare the prior works briefly reviewed above in parallel from multiple aspects.

\section{Task Definition and Problem Formulation}
\label{sec:problem_formulation}

As the pioneering work to propose the task of automatic related work generation, \citet{hoang-kan-2010-towards} envision the related work summarization task that assists in creating a draft of the related work section. They frame it as a topic-biased, multi-document summarization problem that takes a target scientific document for which a related work section needs to be drafted and a list of cited papers as the inputs, and the output goal is to create a related work section that contextually describes the related prior works in relationship to the scientific document at hand. They identify the main steps of related work generation: (1) Finding relevant documents; (2) Identifying the salient aspects of these documents in relation to the current work worth summarizing; (3) Generating the topic-biased summary.

This general task definition becomes widely acknowledged by the successive works approach this problem. However, since this problem definition is derived from the practical need of generating related work sections automatically, it is a loose definition without having strict constraints. For example, there is no clear restriction of the format of the related work section, such as length, writing style, etc. There is also no clear definition of how the target paper or the set of cited papers look like. As a result, each author may interpret the task definition in their way, and they may be all valid. 

In addition to the different understanding of the task definition, each work may have different scopes of the task, such as only addressing part of the task steps or only considering part of the input papers. In this section, we compare each prior work's specific task definition, and the input and output of their systems. Table \ref{tab:task_formulation_extractive} summarizes the inputs to the extractive systems, and Table \ref{tab:task_formulation_abstractive} summarizes the inputs and outputs to the abstractive systems.

\subsection{Extractive Related Work Generation}
\citet{hoang-kan-2010-towards}'s task definition of related work generation lay a foundation of how the successive works approach this problem. Instead of addressing all steps of related work generation that they identified, they only focus on generating the full topic-biased summary given an existing topic hierarchy tree of the target related work and each cited full paper. Their system's output is also simplified to a sequence of extracted sentences, without having separation of paragraphs. Retrospectively, they are the only work that requires a pre-built topic tree as the input. 

\citet{hu2014automatic} inherit the task definition of related work generation from \citet{hoang-kan-2010-towards}, but they consider the full related work generation process by generating related work sections given a list of cited papers and the rest of the target paper. They take the abstract and introduction sections of the target paper and the abstract, introduction, related work, and conclusion sections of the cited papers as their inputs. Their outputs are also a sequence of extracted sentences. 

Similarly, \citet{chen2019automatic} take the title, abstract, introduction, and conclusion of the target paper and each of the cited papers as their inputs for keyword extraction. They use the papers that co-cite the cited papers of the target related work as the extraction sources. 

Later, \citet{wang2019toc} consider full papers of the target paper excluding the related work section, and full papers of the cited papers as well as the citation sentences that co-cite the cited papers, so that they can leverage CTS for their generation step. 

\citet{deng2021automatic} directly extract sentences from the abstract and conclusion section of the cited papers, without directly relying on the citation sentences from training, so that their approach does not suffer from the delay of citing publications for newly published papers, or require the input of citation data.

In general, extractive related work generation approaches take the sentences from the rest of the target paper, the cited papers, and optionally co-citing sentences as the inputs, and concatenate the extracted sentences in a certain order to form a single ``paragraph'' of summary as the generated related work section. As a result, the output summary does not have a division of paragraphs, or any salient structure, which is very different from the actual related work sections commonly seen in NLP papers.

\begin{table*}[t]
\begin{center}
    \begin{tabular}{ | l | l | }
    \hline
    \textbf{Prior Work} & \textbf{Inputs} \\ \hline
    \citet{hoang-kan-2010-towards} & Topic hierarchy tree of the target related work, full cited papers \\  \hline 
    \citet{hu2014automatic} & Target paper (abstract, introduction), \\ 
    & cited papers (abstract, introduction, related work, conclusion) \\ \hline
    \citet{chen2019automatic} &  Title, abstract, introduction, and conclusion for both target \\
    & paper and cited papers; papers that co-cite the cited papers \\ \hline
    \citet{wang2019toc} & Full papers of target paper and cited papers, \\ 
    & citation sentences that co-citing the cited papers\\ \hline
    \citet{deng2021automatic} & Abstract or conclusion sections of the cited papers \\ \hline
    \end{tabular}
    \caption{A summary of the problem formulations of the prior works on extractive related work generation. All of their generation targets are a sequence of extracted sentences.} \label{tab:task_formulation_extractive}
    \vspace{-1em}
\end{center}
\end{table*}

\subsection{Abstractive Related Work Generation}
Because abstractive summarization approaches generate outputs token by token, it's much more challenging than extractive approaches to generate a long sequence of text. As a result, most of the current abstractive related work generation works limit their scope to generating citation sentences, which are the most challenging part of related work generation. These works may even not mention ``related work generation'' as their task, but instead, phrase it as ``citation text generation'' \cite{xing2020automatic, ge-etal-2021-baco} or ``explaining relationships between scientific documents'' \cite{luu-etal-2021-explaining}, but they are essentially a fundamental component of related work generation task.

As a pioneering work abstractive citation sentence generation, \citet{abura2020automatic} work on the abstractive generation of citation sentences given the concatenation of cited paper's title and abstract. In parallel, \citet{xing2020automatic}'s model takes 3 sentences each before and after the target citation sentence, and the cited paper's abstract as the inputs. Their generation target is a citation sentence with a single citation, which can be either explicit or implicit. Their problem formulation is a great simplification of citation text generation. Their system can only take a single cited paper as inputs, which makes the generated citation sentence be only about a single cited work.

\citet{ge-etal-2021-baco} employ multiple heterogeneous inputs, and use multi-task learning approach to learn from multiple targets. Their inputs include a citation network, which indicates which paper cites which paper; context sentences defined similarly as \citet{xing2020automatic}; and the cited paper's abstract. Their outputs are the citation sentence, the corresponding citation function of the citation sentence, and a label sequence that highlights salient sentences in the cited abstract. Although they do not limit to generating sentences with a single citation, since they allow only one input cited abstract, their model is essentially only capable of generating citation sentences derived from a single cited paper.

\citet{luu-etal-2021-explaining} tried two types of inputs in addition to the target paper's introduction section: the abstract of the cited paper or the named entities and terms extracted from the cited paper. Their generation target is the citation sentence.

Contrasting to the setting of all abstractive approaches above, \citet{chen-etal-2021-capturing} is the only work that generates the entire paragraph with two or more citations, which is closer to the related work generation task in the real-life setting.

In general, the state-of-the-art abstractive related work generation approaches limit their scope to much shorter inputs and outputs than the actual available text resources, by using the abstracts of the cited papers as the proxy of the entire paper. Most works \cite{xing2020automatic, ge-etal-2021-baco, luu-etal-2021-explaining} simplify the citation text generation task to a biased summarization of a single cited paper, which is far from the full version of automatic related work generation. 

\begin{table*}[t]
\begin{center}
    \begin{tabular}{ | l | l | l | }
    \hline
    \textbf{Prior Work} & \textbf{Inputs} & \textbf{Target} \\ \hline
    \citet{abura2020automatic} & Cited title, abstract & Citation sentence w/ single \\
    & & reference \\ \hline
    \citet{xing2020automatic} & Context sentences, & Citation sentence w/ single \\  
    & single cited abstract & reference \\ \hline
    \citet{ge-etal-2021-baco} & Citation network, single cited  & Citation sentence, citation \\
    & abstract, context sentences &  function, salient sentence \\ 
    & &  in cited abstracts \\ \hline
    \citet{luu-etal-2021-explaining} & Intro of the citing paper, & Citation sentence w/ single reference \\  & named entities of the cited papers & \\ \hline
    \citet{chen-etal-2021-capturing} & Cited abstracts & A paragraph w/ 2+ citations \\ \hline
    \end{tabular}
    \caption{A summary of the problem formulations of the prior works on neural network-based related work generation. ``Context'' refers to those sentences or paragraphs around the target citation sentences.} \label{tab:task_formulation_abstractive}
    \vspace{-1em}
\end{center}
\end{table*}

\section{Datasets and Preprocessing}
\label{sec:dataset}
Overall, the dataset for related work generation is inexpensive to obtain, since the target related work sections are part of the academic papers that are massively available online. A large portion of the online academic papers is open access, thus allowing researchers to scrape them automatically without paying for the access fee. As a result, there are abundant training materials for related work generation.

\subsection{Commonly Used Corpora}

\subsubsection{ACL Anthology Network}
The ACL Anthology Network (AAN) \cite{radev2013acl}, initiated by \citet{bird2008acl}, collects information regarding all of the papers included in the many Association for Computational Linguistics (ACL) venues. It consists of ``a comprehensive manually curated networked database of citations, collaborations, and summaries in the field of Computational Linguistics'' \cite{radev2013acl} as well as papers published by the ACL. It has been used by many of the surveyed papers.

\subsubsection{S2ORC}
The Semantic Scholar Open Research Corpus (S2ORC) \cite{lo-wang-2020-s2orc} is a recent large corpus of 81.1M English-language academic papers spanning many academic disciplines. The corpus consists of rich metadata, paper abstracts, resolved bibliographic references, as well as structured full text for 8.1 M open access papers. For those full texts, they are annotated with automatically detected inline mentions of citations, figures, and tables. S2ORC is constructed using data from the Semantic Scholar literature corpus \cite{ammar2018construction}, which derives papers from numerous sources: obtained directly from publishers, from resources such as Microsoft Academic Graph (MAG) \cite{shen2018web}, from various archives such as arXiv or PubMed, or crawled from the open Internet in the format of PDFs or LaTeX. Thus S2ORC consists of a large number of parsed paper texts from various fields of study, including medicine, biology, physics, mathematics, computer science, chemistry, psychology, engineering, etc. in decreasing order. Since each inline citation, figure and table are annotated, S2ORC greatly reduces the workload of researchers who work on scientific document processing, by saving the hassle of collecting and pre-processing the raw PDF files. As \citet{chen-etal-2021-capturing, luu-etal-2021-explaining} use S2ORC for their studies. we believe more and more future works will develop their work based on S2ORC.

\subsection{Overview of Datasets Used in Related Work Generation} 
Table \ref{tab:datasets} summarizes the datasets of the surveyed studies collect. Because of the professionality of academic papers, reading papers from unfamiliar domains is challenging. Therefore, prior works mostly focus on the computational linguistic domain that NLP researchers are naturally familiar with. 

\subsubsection{Extractive Approaches} \hfill \break

\citet{hoang-kan-2010-towards} create a related work dataset called RWSData, which includes 20 articles from prestigious NLP and IR venues, namely SIGIR, ACL, NAACL, EMNLP, and COLING. They further manually annotate each of the related work sections collected with a topic tree. Each article's data consist of the citing related work summary, the collection of the input research articles that were cited, and a manually-constructed topic tree.

\citet{hu2014automatic} build a corpus that contains academic papers and their corresponding cited papers from the ACL Anthology \footnote{\url{http://aclweb.org/anthology/}}. They remove the papers that contain related work sections with a very short length, and randomly select 1050 target papers. They extract the texts from PDF files using PDFlib \footnote{\url{http://www.pdflib.com/}} and detect the physical structures of paragraphs, subsections and sections using ParsCit \footnote{\url{http://aye.comp.nus.edu.sg/parsCit/}}. The related work sections are directly extracted as the gold summaries. The cited papers are downloaded from ACL Anthology if available, otherwise, they are downloaded using Google Scholar. They discard cited Ph.D. theses and books.

\citet{chen2019automatic} build a dataset for 25 papers from the ACL Anthology and IJCAI. They extract a maximum of 10 citation sentences from the papers that co-citing each cited paper of the target related work. For this step, they use Google Scholar and Baidu Scholar to retrieve the co-citing papers and use the pdfbox library \footnote{\url{http://pdfbox.apache.org/}} to extract the texts from PDF files and use StanfordCoreNLP \cite{manning2014stanford} toolkit to split the texts into sentences. Then they use a rule-based approach to extract citation sentences from the co-citing papers.

\citet{wang2019toc} train their CTS identification model using CL-SciSumm2018 \footnote{\url{http://wing.comp.nus.edu.sg/ cl-scisumm2018/}} dataset. They collect 50 papers with more than 10 reference papers from NLP conferences (ACL, EMNLP, NAACL, COLING) as the target papers with a time span ranging from 2006 to 2017. The texts are extracted using pdfminer \footnote{\url{https://pypi.org/project/pdfminer/}} from the paper PDFs. They use specifically designed regular expressions to identify the list of references semi-automatically, then use Google Scholar to retrieve the reference papers. For the citation sentence collection, they download the top 20 citing papers of each reference from Google Scholar and extract the sentences using extraction rules. Each reference paper possesses 3 to 20 citations.

\citet{deng2021automatic} derive their dataset automatically from the ScisummNet corpus \cite{yasunaga2019scisummnet}. To automatically annotate salient sentences that should be extracted to form the related work section, they make use of the citation sentences of a paper in the corpus to produce a gold label of whether a sentence in the paper is salient. Their annotating algorithm is similar to \citet{nallapati2017summarunner}: they repeatedly select sentences one-by-one from the cited papers' abstract or conclusion sections to maximize the ROUGE score \cite{lin2004rouge} between the selected set of sentences and the concatenation of all of the citation sentences in the target paper. They label all selected sentences in the cited papers' abstract and conclusion sections as salient and other sentences as not salient. In this way, they are able to quickly collect a dataset with 11954 training samples with 3541 positive ones.

\begin{table*}[t]
\begin{minipage}{\textwidth}
\begin{center}
    \begin{tabular}{ | l | l | l | l |}
    \hline
    \textbf{Prior Work} & \textbf{Source Domain} & \textbf{Size} & \textbf{Dataset} \\ \hline
    \citet{hoang-kan-2010-towards} & Papers from NLP and IR, & 20 papers & RWSData \footnote{\url{http://wing.comp.nus. edu.sg/downloads/rwsdata}} \\ 
    & manually curated topic tree & & \\ \hline 
    \citet{hu2014automatic} & ACL Anthology & 1050 papers & N/A \\ \hline
    \citet{chen2019automatic} & ACL Anthology \& IJCAI & 25 papers & RWS-Cit \footnote{\url{https://github.com/jingqiangchen/RWS-Cit}} \\\hline
    \citet{wang2019toc} & NLP conferences & 50 papers & NudtRwG \footnote{\url{https://github.com/ NudtRwG/NudtRwG-Dataset}} \\ \hline
    \citet{deng2021automatic} & ScisummNet (ACL) & 11954 examples & N/A \\ \hline
    \hline
    \citet{abura2020automatic} & ScisummNet (ACL) & 940 + 15574 pairs & N/A \\
    \hline
    \citet{xing2020automatic} & ACL Anthology Network & 1k + 85k examples & Available \footnote{\url{https://github.com/XingXinyu96/citation_generation}} \\ \hline
    \citet{ge-etal-2021-baco} & ACL Anthology Network & 1.2k + 84k examples & N/A \\ \hline
    \citet{chen-etal-2021-capturing} & S2ORC (Multi-domain), Delve (CS) & 150k, 80k examples & N/A \\ \hline
    \citet{luu-etal-2021-explaining} & S2ORC (CS) & 622k citations & Extraction \\
    & & from 154k papers & from S2ORC \footnote{\url{https://github.com/Kel-Lu/SciGen/tree/master/data_processing}} \\
    \hline
    \end{tabular}
    \caption{A summary of the datasets of the prior works on related work generation.} \label{tab:datasets}
    \vspace{-1em}
\end{center}
\end{minipage}
\end{table*}

\subsubsection{Abstractive Approaches} \hfill \break
\citet{abura2020automatic} collect 940 pairs of <title, abstract, citation sentence> pairs from ScisummNet \cite{yasunaga2019scisummnet}, and further collect 15574 pairs from MAG \cite{shen2018web} and Open Academic Graph (OAG), which is a large knowledge graph unifying two billion records from MAG and AMiner \cite{tang2016aminer}. They use a filter based on \citet{teufel1999argumentative}'s first pronoun and presentation nouns gazetteers to filter the sentences in the abstracts.

\citet{xing2020automatic} build their citation text generation dataset based on the 2014 version of ACL Anthology Network (AAN) corpus \cite{radev2013acl}, which is a collection of papers from the Computational Linguistics journal, and proceedings from ACL conferences and workshops. They extract 86052 explicit citations from 16675 papers. They manually annotate 1000 citation texts, each of which consists of an explicit citation sentence with three surrounding sentences before and after that are possibly implicit citation sentences. They also use their trained model to annotate 85052 automatically extracted citation texts. 

\citet{ge-etal-2021-baco} follow \citet{xing2020automatic} to extract citation sentences using regular expressions from AAN \cite{radev2013acl}. They manually annotate the citation functions on 1200 citation sentences and train a SciBERT-based \cite{beltagy2019scibert} model to automatically label the rest of the dataset to build a large dataset.

\citet{chen-etal-2021-capturing} collect their dataset from S2ORC \cite{lo-wang-2020-s2orc}, which consists of papers in multiple domains (physics, math, computer science, etc.) and Delve \cite{akujuobi2017delve}, which consists of computer science papers. Their dataset size is much larger than the prior works. They collect 150k and 80k abstract-related work paragraph pairs for S2ORC and Delve respectively.

Similarly, \citet{luu-etal-2021-explaining} derive their dataset for citation text generation from S2ORC \cite{lo-wang-2020-s2orc}. They use 154k connected computer science articles, from which they extract 622k citation sentences with a single reference that links back to other documents in S2ORC. Specifically, sentences with more than one reference are omitted.

\subsection{Discussion}
All of the datasets used in the related work generation studies concentrate on the computer science domain, most of which are purely on computational linguistics. ACL papers and S2ORC dataset \cite{lo-wang-2020-s2orc} are the most popular original sources to extract the related work sections from. In the early works whose dataset only consist of less than 50 papers, the text extraction process involves more manual operation. However, the size of the datasets studied in abstractive approaches is significantly larger than the extractive approaches. This is because abstractive approaches are all based on neural networks, which require a much larger size of datasets. \citet{xing2020automatic, ge-etal-2021-baco} both take the approach of using a model trained on a small manually curated dataset to label a large-scale unlabeled dataset to obtain a large labeled dataset. Ironically, every single work discussed collected their own dataset from scratch, despite the fact that many of the works release the datasets they collected. The reason that there is no widely recognized off-the-shelf dataset is probably that different models require different sizes of datasets, with a different format of input requirements that cannot be adapted from previously released datasets.

\section{Approaches}
\label{sec:approaches}
As we indicate before, the existing works can be largely divided into extractive and abstractive related work generation. Most earlier existing works are all extractive, while later works since 2020 are all in abstractive fashion using neural networks, which is also consistent with the evolution of general domain summarization studies. Table \ref{tab:approaches} summarizes the approaches of the surveyed works.

\begin{table*}[t]
\begin{center}
    \begin{tabular}{ | l | l | }
    \hline
    \textbf{Prior Work} & \textbf{Approaches} \\ \hline
    \citet{hoang-kan-2010-towards} & Heuristic approach to generate general and specific content\\ 
    & separately given a topic tree \\ \hline 
    \citet{hu2014automatic} & PLSA for topic modeling, SVR for sentence importance \\ 
    & score, and global optimization for sentence selection \\ \hline
    \citet{chen2019automatic} & Considering papers co-cite the cited papers; \\
    & Representing graph for relationship modeling of papers, then finding \\
    & sentence nodes that cover the minimum Steiner tree of the graph \\ \hline
    \citet{wang2019toc} & Leveraging both topic model and cited text spans \\ \hline 
    \citet{deng2021automatic} & BERT-based sentence extraction \& reordering \\ \hline
    \hline
    \citet{abura2020automatic} & Applying PTGen and Transformer \\ \hline
    \citet{xing2020automatic} & Manual annotation + automatic annotation of citation sentences; \\  
    & PTGEN-Cross based on cross-attention mechanism \\ \hline
    \citet{ge-etal-2021-baco} & Citation network as auxiliary input; citation function \& salient \\ 
    & sentences in cited papers as auxiliary output; multi-task learning \\ \hline
    \citet{luu-etal-2021-explaining} & SciGPT2; IE-Extracted Term Lists; ranking based on entity matching \\ \hline
    \citet{chen-etal-2021-capturing} & Transformer-based hierarchical encoder; relationship modeling module\\
    \hline
    \end{tabular}
    \caption{A summary of the approaches of the prior works.} \label{tab:approaches}
    \vspace{-1em}
\end{center}
\end{table*}

\subsection{Extractive Related Work Generation}
\subsubsection{Heuristic-based Approach} \hfill \break
As the first work about related work generation, \citet{hoang-kan-2010-towards}'s work is published in 2010, before the era of the prevalence of neural networks. Their related work summarizer (ReWoS) is a largely heuristic system, with two independent different strategies for general content summarization (GCSumm) and specific content summarization (SCSumm), which can be easily topologically mapped to the topic tree that they require as the input. The general content is described in tree-internal nodes, whereas leaf nodes contribute detailed specifics. They use keywords to exclude input sentences or classify sentences to GCSumm or SCSumm options. 

For GCSumm, \citet{hoang-kan-2010-towards} distinguish informative and indicative sentences. Informative sentences give details, such as definitions, purpose, or application on a specific aspect of the problem (\textit{``Text classification is a task that assigns a certain number of predefined labels for a given text.''}). In contrast, indicative sentences are simpler, inserted to make the topic transition explicit and rhetorically sound (\textit{``Many previous studies have studied monolingual text classification.''}). They generate indicative sentences with templates and use Term Frequency $\times$ Inverse Sentence Frequency (TF $\times$ ISF) scores \cite{otterbacher2005using} to extract sentences from each cited article as informative sentences.

SCSumm uses a similar extractive approach to select sentences. In addition, they choose nearby sentences within a contextual window after an agent-based sentence to represent a topic. To address the issue that the presence of one or more current, ancestor, and sibling nodes may affect the final score from the computation of sentence importance, they assign different weights to the candidate sentences based on the positions of their corresponding nodes.

Their final summary output is a sequence of extracted sentences, with post-processing to replace pronouns such as ``we'' to the formal citation marks, such as ``(Wu and Oard, 2008)''.

\subsubsection{Extracting Sentences Based on Topic Modeling} \hfill \break
\citet{hu2014automatic} inherit the task setting of \citet{hoang-kan-2010-towards} but propose a solution that does not require a manually curated topic tree. They first use the PLSA \cite{hofmann1999probabilistic}, which is a topic model to compute the topic information for each sentence in the target paper and cited papers. In parallel, they train two separate SVR models to assess the sentence importance by computing the similarity score between each sentence in the target paper or cited papers and the sentences in the target related work section. They create a series of features to perform the regression. Finally, the related work generation process is formulated as a global optimization problem by optimizing an objective function based on the topic information and the sentence importance score.  

\citet{chen2019automatic} models the relationship between two documents with the representing graph. They use $\chi^{2}$ statics as the feature selection method to select discriminative features to construct the feature graph that represents two documents. The discriminative features obtained from the feature selection step represent the differences between the two documents. The discriminative features may not be well connected in the feature graph, so they connect the discriminative nodes (sentences) and the nodes shared by two documents. They formulate the discriminative features connecting problem as the minimum Steiner tree problem to connect the features in a meaningful way. Due to the NP-hardness \cite{fraer1999minimum}, they approximate the minimum Steiner tree problem with Floyd-Warshall algorithm \cite{floyd1962algorithm} to find the shortest path of the root node to the terminal nodes and then merge the paths. Finally, they select the minimum set of subgraphs that cover the Steiner tree, which is approximated with a greedy algorithm. This is essentially extracting the sentences from the papers that co-cite the same reference papers with the target paper.

\subsubsection{Identifying Cited Text Spans} \hfill \break
In addition to a topic model implemented by QueryTopicSum, \citet{wang2019toc} also explicitly extract Cited Text Spans (CTS), which are the matched text spans in the cited paper that are most related to a given citation. 

\citet{wang2019toc}'s topic model, QueryTopicSum is inspired by TopicSum \cite{haghighi2009exploring}, which is designed for general multi-document summarization. QueryTopicSum constructs a target-reference distribution for each cited paper, in addition to TopicSum-like background distribution and document-specific distribution, in order to capture the relationship between the target paper and the reference papers.

In addition to the target paper excluding the related work and the cited papers, \citet{wang2019toc} also consider citation sentences from the papers that are co-citing the cited papers. Hence, CTS can be viewed as important sentences annotated by the community. They use a two-layered ensemble model to extract CTS as a classification task. In the first layer, they use Random Forest classifiers based on features including Jaccard similarity, BM25 \cite{jones2000probabilistic} similarity, Term frequency-inverse document frequency (TF-IDF) similarity, section similarity, and two word embedding similarities from \citet{cohan2018scientific}. They train 25 Random Forest classifiers on CL-SciSumm2018 \cite{jaidka2019cl} dataset for CTS classification. Their second layer takes the BM25, TF-IDF and the output of the first ensemble layer to perform another vote for predicting CTS.

Finally, a greedy algorithm selects the candidate sentences to generate the related work section. The algorithm selects the sentences that minimize the KL divergence between the true distribution of the related work sentences and the approximating distribution during each iteration until reaching a limit.

\subsubsection{Sentence Extraction and Reordering} \hfill \break
\citet{deng2021automatic}'s approach consists of two steps. They first use an ensemble sentence classification model to extract salient sentences from the cited papers' abstract and conclusion sections. Next, they reorder the extracted sentences as a next sentence prediction (NSP) task. 

Their ensemble sentence classification model consists of two parts. The first one is a BERT model \cite{devlin2018bert} that computes the probability of salience sentences by sentence. The second part is a bag-of-words sentence matching model, which simply checks whether the input sentence contains any of the curated keywords, such as ``novel'', ``propose'', and ``improve''. Then they use a step function to merge the scores obtained from these two models and only choose top-3 salient sentences for the sentence reordering.

In the sentence reordering problem, the order of each sentence is supposed to depend on the order of all its previous sentences, which is hard to compute directly. Instead, they make a Markov assumption similar to \citet{chen2016neural},
by assuming that each sentence's probability only depends on its most recent sentence. In this way, the reordering problem is simplified to the NSP task, which generates a probabilistic value indicating whether the first sentence in the input is followed by the second one in the source document. For this NSP task, they use a MobileBERT \cite{sun2020mobilebert} model, which is a compact task-agnostic BERT \cite{devlin2018bert} that runs more than 5 times faster than BERT-base while still achieving comparable results. They fine-tune MobileBERT on NSP task on ScisummNet corpus \cite{yasunaga2019scisummnet} with 360k training examples. Finally, the re-ordered sentences are modified with citation tags and proper pronouns to form the final output of the system.

\subsection{Abstractive Related Work Generation}

\subsubsection{Neural Network for Abstractive Citation Sentence Generation} \hfill \break
\citet{abura2020automatic} takes a simple sequence-to-sequence approach: they apply the classic architectures and implementations to citation sentence generation given the title and abstract of the cited paper. They apply the OpenNMT-py \cite{klein2017opennmt} implementation of the Pointer-Generator Network with copy-attention and coverage mechanism \cite{see-etal-2017-get}, and Transformer \cite{vaswani2017attention} to this task.

\subsubsection{Abstractive Citation Text Generation Given Citation Text} \hfill \break

\citet{xing2020automatic} first use regular expressions for matching reference signs in explicit citation sentences. They define the context sentences of each explicit citation sentence as the three sentences before and after the explicit citation sentence. They refer the explicit citation sentence and their context sentences together as citation texts. Then annotators are asked to find out consecutive implicit citation sentences from the context sentences of each explicit citation sentence. The majority vote is used to resolve the disagreement among annotators. They collect 1000 manually annotated citation texts, each of which contains an explicit citation sentence and its associated implicit citations. They only consider explicit citation sentences with only one citation.

Next, \citet{xing2020automatic} train a BERT \cite{devlin2018bert} model to extract citation sentences given each unlabeled citation text, by formulating it as a binary sequence labeling problem. They also concatenate the abstracts of the cited paper to the citation texts in their inputs. As a result, they obtain an additional 85k automatically extracted citation texts.

For citation generation, \citet{xing2020automatic} propose the PTGEN-Cross model, a variation of the Pointer-Generator Network \cite{see-etal-2017-get} with a cross-attention mechanism so that the model takes both the citation context and the cited paper's abstract as inputs and outputs the citation sentence. Their experiments verify the effectiveness of the cross-attention mechanism.

\subsubsection{Using External Information to Improve Citation Sentence Generation} \hfill \break
\citet{ge-etal-2021-baco} inherit the citation sentence generation task setting from \citet{xing2020automatic} and improve their performance by introducing external information including background knowledge in the form of citation networks, and content from both citing and cited papers. From the citing papers' side, in addition to \citet{xing2020automatic}'s approach of labeling citation sentences in AAN, they label each citation sentence with citation function labels in \{Positive, Negative, Neutral, Mixed\}. They further fine-tune a SciBERT \cite{beltagy2019scibert} for automatically tagging citation function to the unlabeled citation sentences. From the cited papers' side, they adopt a ROUGE-based approximation approach \cite{yasunaga2017graph} to retrieve salient sentences in each cited paper's abstracts that are highly related to the citation sentence.

Given these additional labels, \citet{ge-etal-2021-baco} propose their end-to-end model that extends the Pointer-Generator Network \cite{see-etal-2017-get} and takes the citing sentence's context, cited paper's abstract as well as the citation network information as inputs and outputs the citation sentence, the citation function label and the salience estimation of the sentences in the cited paper's abstract. Specifically, they use a Graph Attention Network (GAT) \cite{Velickovic2018GraphAN} to encode the citation network, and a hierarchical RNN-based encoder to encode the citation context and the cited abstract. Their decoder is similar to \citet{see-etal-2017-get}'s pointer-generator network. They train their model jointly on three types of outputs using multi-task learning \cite{caruana1997multitask}.

\subsubsection{Relation-Aware Related Work Generation} \hfill \break
\citet{chen-etal-2021-capturing}'s RRG model has three main parts: (1) The hierarchical encoder implemented by a Transformer \cite{vaswani2017attention} encodes multiple input cited abstracts. They implement two types of positional encoding to encode document index and word index separately. (2) The relationship modeling part fuses encoded cited abstract representations to relation-aware document encoding and a polished relation graph. They customize a stacked Transformer-block-like \cite{vaswani2017attention} architecture, and each block contains a relation graph updater that consists of a multi-head attention mechanism \cite{vaswani2017attention} and an LSTM-like \cite{hochreiter1997long} gate. (3) Finally a Pointer-Generator Network-like \cite{see-etal-2017-get} module is used for generating related work.

\subsubsection{Fine-tuning Pretraiend Language Model} \hfill \break
\citet{luu-etal-2021-explaining} experiment with two approaches to obtain the explanation of the relationship between two scientific papers, which uses the citation as an approximated answer. In the first approach, they pretrain a GPT2 \cite{radford2019language} in the science domain just like SciBERT \cite{beltagy2019scibert} to produce SciGPT2, which is further trained on their citation text dataset based on S2ORC to produce SciGen as their model for generating citation sentences. They use SciGen to generate a citation sentence given the introduction of the citing paper and the abstract of the cited paper. The second approach is an IR-based approach, which retrieves the citation sentence from the citing-cited paper pair closest to the target citing-cited paper pair. The similarity is computed by a weighted cosine similarity over the SciBERT embedding \cite{beltagy2019scibert} of the entire paper abstract. 
They further try information extraction techniques to extract named entities or terms from the cited papers to enrich the paper representation. They try both named entity extraction and terms with top TF-IDF scores. To combat hallucinations and promote factual accuracy, they use mean reciprocal rank to rank the candidate answers based on the entity matching between the candidate sentences and the cited document. 

\subsection{Summary}
Since each academic paper requires some novel contributions, as the main part of an experimental paper, the approach section is usually full of novelty. Although there are only a limited number of prior works on related work generation, we can obtain some insights on the helpful techniques to improve the system performance. The existing extractive systems typically focus on selecting existing sentences that thoroughly but also concisely cover the main ideas of the prior works. Topic modeling and scoring each candidate sentence seem to be necessary components, as they are addressed by all four extractive works \cite{hoang-kan-2010-towards, hu2014automatic, chen2019automatic, wang2019toc}. Taking more inputs into considerations is also an important direction to explore, such as related work sections that co-cite the cited papers \cite{chen2019automatic}, CTS \cite{wang2019toc} or cited sentence salience \cite{ge-etal-2021-baco}, citation network and citation function \cite{ge-etal-2021-baco}. For abstractive approaches, it's crucial to design an appropriate neural network architecture to encode and fully leverage the input information, such as cross-attention mechanism \cite{xing2020automatic}, hierarchical encoders \cite{ge-etal-2021-baco, chen-etal-2021-capturing}. Moreover, learning techniques such as transfer learning from general domain to scientific domain via language model fine-tuning \cite{luu-etal-2021-explaining} and multi-task learning \cite{ge-etal-2021-baco} are helpful techniques that can be easily inherited by future work. Finally, it is also one of the most crucial topics to combat hallucinations and promote factual accuracy for abstractive text generation tasks. \citet{luu-etal-2021-explaining} explore term-based approaches to enforce the factually, but there is still a lot of room to explore other effective approaches.

\section{Results and Evaluations}
\label{sec:evaluations}
Evaluation of the generated text is a crucial step for automatic summarization. In this section, we review the performance of the proposed systems, and how those systems are evaluated and compared to their baselines.

\subsection{Evaluation methods}
 \citet{gambhir2017recent} categorize performance evaluation methods of text generation systems into two types: extrinsic evaluation and intrinsic evaluation. Extrinsic evaluation determines the summary's quality based on how it affects other tasks (text classification, information retrieval, question answering). In other words, a good summary should provide help to other downstream tasks. This can be measured by time-to-completion, success rate and decision-making accuracy, etc. \cite{altmami2020automatic}. On the other hand, intrinsic evaluation determines the summary quality based on the coverage between the automatically generated summary and the human-generated summary. Quality or informativeness are the two important aspects based on which a summary is evaluated \cite{gambhir2017recent}. Usually, the informativeness of a candidate summary is evaluated by comparing it with a reference summary generated by a human. This is usually measured by automatic evaluation metrics such as accuracy, precision, recall, F-measure, ROUGE \cite{lin2004rouge}, BLEU \cite{papineni2002bleu} etc. Although automatic evaluations are fast and inexpensive, they usually examine the word-level coverage without considering the semantic similarity. As a result, examples with high ROUGE or BLEU scores may not necessarily be the best ones. The other intrinsic evaluation is to assess the quality of the generated text by a human from the aspects of grammaticality, succinctness, and coherence, etc. The problem with this technique is that there is no single ``ideal'' target \cite{el2011exploring}, as one can generate multiple different valid summaries for the same document, and moreover, the same summary can be expressed using distinct word choices. Furthermore, unique but valid summaries can be created with respect to different goals depending on the summarization task itself \cite{lloret2012text}.

\hfill \break
Next, we briefly describe the evaluation approach frequently   used in the surveyed literature. 

\subsubsection{ROUGE}
ROUGE \cite{lin2004rouge} stands for Recall-Oriented Understudy for Gisting Evaluation. This is a fully automated evaluation. ROUGE is a software package and containing a set of metrics that is popular and widely used for evaluating automatic summarization. These metrics compare an automatically generated summary with a human-generated reference or reference summary by counting the number of overlaps between the candidate summaries and the reference. All prior works for related work generation use ROUGE \cite{hoang-kan-2010-towards, hu2014automatic, chen2019automatic, wang2019toc, xing2020automatic, ge-etal-2021-baco, chen-etal-2021-capturing, luu-etal-2021-explaining}. There are several ROUGE measures, including ROUGE-N (n-gram co-occurrence statistics, e.g. ROUGE-1, ROUGE-2), ROUGE-L (longest common subsequence), ROUGE-W (weighted longest common subsequence), and ROUGE-S (skip-bigram co-occurrence statistics). Although ROUGE is fast to compute, it cannot substitute human evaluation. Since there is no single best reference, a good summary may be penalized for including relevant phrases that are not in the reference. In addition, since the calculation is overlap-based, ROUGE does not properly measure the semantic similarity between the candidate summary and the reference. Therefore, some summaries would be penalized even they contain segments semantically similar to the reference summary. 

\begin{table*}[t]
\begin{center}
    \begin{tabular}{ | l | l | l | l | }
    \hline
    \textbf{Prior Work} & \textbf{Baselines} & \textbf{Automatic} & \textbf{Human Evaluation} \\ \hline
    \citet{hoang-kan-2010-towards} & LEAD, MEAD & ROUGE recall & Correctness, novelty,  \\ 
    & & (1, 2, S4, SU4) & fluency, usefulness \\\hline 
    \citet{hu2014automatic} & MEAD, LexRank & ROUGE F1 & Correctness, readability,  \\ 
    && (1, 2, SU4) & usefulness\\\hline 
    \citet{chen2019automatic} & MEAD, LexRank, RoWoS & ROUGE F1 (1, 2) & N/A \\ \hline
    \citet{wang2019toc} & LexRank, SumBasic, & ROUGE recall  & N/A \\
    & JS-Gen, TopicSum & \& F1 (1, 2, SU4) & \\ \hline 
    \citet{deng2021automatic} & MEAD & ROUGE precision, & informativeness, \\
    & & recall, F1 & fluency, succinctness \\
    & & (1, 2, L) & \\ \hline
    \hline
    \citet{abura2020automatic} & MEAD, TextRank, SUMMA, & ROUGE precision, & N/A \\
    & SEQ$^3$ & recall, F1 & \\
    & & (1, 2, L, SU4) & \\ \hline
    \citet{xing2020automatic} & RandomSen, MaxSimSen, & ROUGE F1 & Readability, Content, \\ 
    & EXT-Oracle, COPY-CIT, & (1, 2, L) & Coherence, Overall \\
    & PTGEN &&\\ \hline
    \citet{ge-etal-2021-baco}& LexRank, TextRank, &ROUGE F1& Fluency, relevance,\\
    &EXT-Oracle, PTGEN, &(1, 2, L)&coherence, overall\\
    &PTGEN-Cross&&\\ \hline
    \citet{chen-etal-2021-capturing} & LEAD, TextRank, & ROUGE F1 & QA, informativeness, \\
    & BertSumEXT, MGSum-ext, & (1, 2, L) & coherence, succinctness \\
    & PTGen+Cov, TransformerABS, & & \\
    & BertSumAbs, MGSum-abs, GS & & \\ \hline
    \citet{luu-etal-2021-explaining} & N/A & BLEU, ROUGE-L & Correct, Specific \\ \hline
    \end{tabular}
    \caption{A summary of the evaluation methods of the prior works.} \label{tab:evaluation}
    \vspace{-1em}
\end{center}
\end{table*}

\begin{table*}[t]
\begin{center}
    \begin{tabular}{ | l | l | l | }
    \hline
    \textbf{perspective} & \textbf{Definition} & \textbf{Used By} \\ \hline
    Fluency, & Does the summary's exposition flow well, & \cite{hoang-kan-2010-towards, hu2014automatic, deng2021automatic}, \\ 
    Readability & in terms of syntax as well as discourse? & \cite{xing2020automatic, ge-etal-2021-baco, chen-etal-2021-capturing} \\
    \hline
    Correctness & Is the summary content relevant to (express the factual & \cite{hoang-kan-2010-towards, hu2014automatic, luu-etal-2021-explaining} \\ 
    & relationship with) the hierarchical topics/cited papers given? & \\ \hline
    Novelty & Does the summary introduce novel information that is & \cite{hoang-kan-2010-towards}\\
    & significant in comparison with the human created summary? & \\ \hline
    Usefulness & Is the summary useful in supporting the researchers to & \cite{hoang-kan-2010-towards, hu2014automatic} \\
    & quickly grasp the related works given hierarchical topics? & \\ \hline
    Content, & Whether the citation text is relevant to the cited paper's & \cite{xing2020automatic, ge-etal-2021-baco} \\ 
    Relevance & abstract & \\\hline
    Coherence & Whether the citation text is coherent with  & \cite{xing2020automatic, ge-etal-2021-baco, chen-etal-2021-capturing}\\ 
    & the citing paper's context & \\\hline
    Informativeness & Does the related work convey important facts & \cite{deng2021automatic, chen-etal-2021-capturing} \\
    & about the topic question? & \\ \hline
    Succinctness & Does the related work avoid repetition? & \cite{deng2021automatic, chen-etal-2021-capturing} \\ \hline
    Overall & Overall quality & \cite{xing2020automatic, ge-etal-2021-baco}\\\hline
    QA & Retain the key information? & \cite{chen-etal-2021-capturing} \\ \hline
    Specific & Whether the explanation describes a specific relationship & \cite{luu-etal-2021-explaining} \\
    & between the two works & \\ \hline
    \end{tabular}
    \caption{A summary of the perspectives for human evaluation.} \label{tab:human_evaluation}
    \vspace{-1em}
\end{center}
\end{table*}

\hfill \break
Next, we review how each of the prior work compare their work against their baselines, and how they perform automatic and human evaluations. Table \ref{tab:evaluation} summarizes the evaluation methods of the surveyed studies, and Table \ref{tab:human_evaluation} summarizes the perspectives used for human evaluations.

\subsection{Extractive Related Work Generation Performance Evaluation}
\citet{hoang-kan-2010-towards} compare their RoWoS system against two baselines, LEAD \cite{wasson1998using} and MEAD \cite{radev2004centroid}. The LEAD baseline takes the first $n$ sentences, usually title and abstract, as the representation of each cited paper. MEAD is an open-source extractive multi-document summarizer. The baseline implementations are provided by MEAD toolkit \footnote{\url{http://www.summarization.com/mead/}}. They evaluate candidate systems automatically using ROUGE recall scores \cite{lin2004rouge}. Their best model obtains the ROUGE-1, ROUGE-2, ROUGE-S4, and ROUGE-SU4 recall scores of 0.698, 0.183, 0.218, and 0.298 respectively. For the human evaluation, they ask 11 judges to score each summary following the perspectives of correctness, novelty, fluency, and usefulness. Each score is on a 5-point scale of 1 (very poor) to 5 (very good). They only evaluate 10 out of 20 evaluation sets. Each set was graded at least 3 times by 3 different evaluators. The examples were randomized so that evaluators did not know the identities of the systems. Their best model obtained 3.6, 3.4, 3.4, 3.6 on correctness, novelty, fluency, and usefulness respectively. As the first work for related work generation, many of their evaluation settings are kept as a conversion for the subsequent studies.

\citet{hu2014automatic} compare their system with MEAD \cite{radev2004centroid} and LexRank \cite{erkan2004lexrank}. LexRank is a graph-based multi-document summarization system based on a random walk on the similarity graph of sentences inspired by Pagerank \cite{page1999pagerank}. They also compare with the variant baselines without considering the content of the target paper. For the automatic metrics, they use ROUGE scores \cite{lin2004rouge}. Their best model, ARWG obtain the ROUGE-1, ROUGE-2, ROUGE-SU4 F-measure of 0.479, 0.122, 0.186 respectively. For the human evaluation, they evaluate the correctness, readability, and usefulness of the candidate-related work sections on the 5-point scale. They ask three graduate student judges from the computer science field to evaluate 15 random target papers each, with randomizing the identities of the candidate sections. Their ARWG model scores 3.4 for all three perspectives respectively.

\citet{chen2019automatic} choose MEAD \cite{radev2004centroid}, LexRank \cite{erkan2004lexrank} and RoWoS by \citet{hoang-kan-2010-towards} as their baselines. They use the precision, recall and F-measure of ROUGE-1 and ROUGE-2 scores \cite{lin2004rouge} as the automatic metric. Their best model scores 0.428 and 0.148 on ROUGE-1 and ROUGE-2 F-measure respectively. They did not conduct human evaluation.

\citet{wang2019toc} compare their QueryTopicSum against LexRank \cite{erkan2004lexrank}, SumBasic \cite{nenkova2005impact}, JS-Gen \cite{peyrard2016general} and TopicSum \cite{haghighi2009exploring}. SumBasic \cite{nenkova2005impact} is a frequency-based summarizer. JS-Gen \cite{peyrard2016general} presents an optimization framework for extractive multi-document summarization. It optimizes JS divergence with a genetic algorithm. TopicSum \cite{haghighi2009exploring} is a hierarchical LDA stype model and presumes that each workd is generated by a single topic which can be a corpus-wise background distribution over common words, a distribution of document-specific words or a distribution of the core content of a given cluster. They use recall and F1 scores of ROUGE-1, ROUGE-2 and ROUGE-SU4 scores \cite{lin2004rouge}. Their best model, QueryTopicSum with CTS scores 0.444, 0.101, 0.173 on ROUGE-1, ROUGE-2 and ROUGE-SU4 F1 scores respectively. They also did not conduct human evaluation.

\citet{deng2021automatic} only compare their SERGE system against MEAD \cite{radev2004centroid}, and use ROUGE \cite{lin2004rouge} scores for automatic evaluation. Their SERGE system obtains 0.297, 0.052, 0.254 F1 scores on ROUGE-1, ROUGE-2, ROUGE-L respectively. For human evaluation they consider informativeness (the property of conveying useful information \cite{liu2019hierarchical}), fluency and succinctness on 5-point scale. Their SERGE system obtains 4.23, 3.83 and 3.83 on informativeness, fluency and succinctness respectively.

\subsection{Abstractive Related Work Generation Performance Evaluation}
As the first work of abstractive citation text generation, \citet{abura2020automatic} compare the Pointer-Generator Network \cite{see-etal-2017-get} and Transformer \citet{vaswani2017attention} with MEAD \cite{radev2004centroid}, TextRank \cite{mihalcea2004textrank}, SUMMA \cite{saggion2008robust} and SEQ$^3$ \cite{baziotis2019seq}. TextRank is an unsupervised algorithm where sentence importance scores are computed based on eigenvector centrality within weighted-graphs. SUMMA is a Java implementation of scoring functions, and \citet{abura2020automatic} use the centroid scoring functionality to select the most central sentence in a document. SEQ$^3$ is an unsupervised abstractive sequence-to-sequence-tosequence autoencoder. Their best model scores 0.300 and 0.113, 0.206, 0.138 F1 on ROUGE-1, ROUGE-2, ROUGE-L, ROUGE-SU4 on the filtered dataset respectively. They did not perform human evaluation.

\citet{xing2020automatic} compare their PTGEN-Cross model against the following baselines: RandomSen, MaxSimSen, EXT-ORACLE, COPY-CIT and PTGEN \cite{see-etal-2017-get}. RandomSen randomly selects a sentence from the cited abstract. MaxSimSen selects the sentence from the cited abstract with the largest similarity with the target citation context. EXT-ORACLE is the upper bound for extractive models, extracted based on ROUGE score \cite{lin2004rouge}. COPY-CIT randomly copies one citation text from the training set that also cites the cited paper. For the automatic metric, they use the F1 of ROUGE-1, ROUGE-2 and ROUGE-L \cite{lin2004rouge}, which becomes a standard for abstractive approaches. Their best model scores 0.263, 0.075, 0.205 on ROUGE-1, ROUGE-2 and ROUGE-L respectively. For human evaluation, they randomly sample 50 instances from the high-recall test set and employ three graduate students to rate the instances in four aspects: readability, content, coherence and overall quality on the 5-point scale. Note that they also score the gold citation texts, which is novel and inherited by the subsequent approaches. Their PTGEN-Cross model scores 3.79, 2.77, 2.85 and 2.91 on readability, content, coherence and overall respectively.

\citet{ge-etal-2021-baco}'s baseline models include LexRank \cite{erkan2004lexrank}, TextRank \cite{mihalcea2004textrank}, EXT-Oracle \cite{xing2020automatic} for extractive approaches, and PTGEN \cite{see-etal-2017-get} as well as PTGEN-Cross \cite{xing2020automatic} for abstractive approaches. Their BACO model achieves scores of 0.325 (ROUGE-1), 0.097 (ROUGE-2), and 0.249 (ROUGE-L). Ablation studies are also conducted by removing some techniques from BACO, such as background knowledge, salience estimation, and citation function. They inherit \citet{xing2020automatic}'s setting by hiring three graduate students who are fluent in English and familiar with NLP to rate 50 instances with respect to fluency, relevance, coherence, and overall quality on the 5-point scale. Their BACO model scores 3.64, 3.07, 2.77, and 2.95 on 
fluency, relevance, coherence, and overall respectively.

\citet{chen-etal-2021-capturing}'s extractive baselines include (1) LEAD \cite{wasson1998using}, which selects the first sentence of each document as the summary; (2) TextRank \cite{mihalcea2004textrank}; (3) BertSumEXT \cite{liu2019text}, which is an extractive summarization model with BERT \cite{devlin2018bert}; (4) MGSum-ext \cite{jin2020multi}, which is a multi-granularity interaction network for extractive multi-document summarization. Their abstractive baselines include (1) PTGen+Cov \cite{see-etal-2017-get}; (2) TransformerABS, which is an abstractive summarization model based on the Transformer \cite{vaswani2017attention}; (3) BertSumAbs \cite{liu2019text}, which is an abstractive summarization network built on BERT \cite{devlin2018bert}; (4) MGSum-abs \cite{jin2020multi}, which is the abstractive version of multi-granularity interaction network; (5) GS \cite{li2020leveraging}, which is a neural abstractive multi-document summarization model that leverages graphs. \citet{chen-etal-2021-capturing}'s RRG model scores 0.291 (ROUGE-1), 0.049 (ROUGE-2), 0.263 (ROUGE-L) on Delve paragraph generation. They also conduct ablation studies by removing some mechanisms they proposed. For human evaluation, \citet{chen-etal-2021-capturing} study 30 randomly selected instances from Delve dataset with two evaluation studies. The first study quantify how the models retain the key information following a question-answering paradigm \cite{liu2019text}, by writing 67 questions based on the ground truth related work, and judging the answers by correct or incorrect. The second evaluation study assess the informativeness, coherence (also including fluency) and succinctness. Their RRG's scores are 38.8\% correct in QA, and 2.37, 2.16 and 2.10 for informativeness, coherence and succinctness respectively, on the 3-point scale. They conduct both of the human evalutation studies on the Amazon Mechanical Turk platform with 3 responses per hit.

\citet{luu-etal-2021-explaining} compares their SciGen model and IR approach with various types of input contexts. They use BLEU \cite{papineni2002bleu} and ROUGE-L \cite{lin2004rouge} as their automatic evaluation metrics. Their best IE-based SciGen model with ranking scores 0.135 on BLEU and 0.123 on ROUGE-L. They conduct two rounds of human evaluations. They first ask 37 NLP researchers and collect 800 judgments on the \textit{correct} and \text{specific} aspects of the generated text on a yes/no scale, with randomization that also includes the gold instances. For the second evaluation on the sentence-based and IE-based models, two of the authors judge 50 data points from each system on a 3-way scale: correct, too value, and incorrect.

\subsection{Summary}
As Table \ref{tab:evaluation} shows, there are a few baselines base widely used across multiple prior studies: LEAD \cite{wasson1998using}, MEAD \cite{radev2004centroid}, LexRank \cite{erkan2004lexrank} and TexRank \cite{mihalcea2004textrank} for extractive approaches and PTGEN \cite{see-etal-2017-get} for abstractive approaches. All these repetitively used baselines are relatively easy to replicate because they are relatively simple and well-documented. 

Overall, all prior works use ROUGE score \cite{lin2004rouge} as the automatic metric for evaluating the generated related work section. \citet{luu-etal-2021-explaining} additionally consider BLEU \cite{papineni2002bleu}. Although the ROUGE scores are not directly comparable across studies since different studies use different settings, such as the task definition and dataset, it's still clear that extractive approaches yield much higher ROUGE scores than abstractive methods, since the extracted sentences may come from the original cited texts. The ROUGE scores of RRG \cite{chen-etal-2021-capturing} is particularly low compared to other abstractive approaches because their output is the entire paragraph.

Most prior works conduct human evaluation since automatic metrics such as ROUGE scores do not correlate with the actual quality well. There is no solid standard of how to conduct the human evaluation. A general strategy is to sample at least 15 examples and get them rated by three evaluators. There is also no fixed perspectives for human evaluation, but generally, each work evaluates the fluency (readability) of the examples, whether the citation text is coherent with the citing paper's context, and the correctness of the content (relevance) with respect to the cited papers. In general, 5-point scale is most popular, but 3-point scale \cite{chen-etal-2021-capturing} and yes/no judgement \cite{luu-etal-2021-explaining} is also employed.

\section{Future Work}
\label{sec:future_work}
Because an experimental paper can only cover a limited amount of experiments, the authors usually state research directions that they are interested in but not yet accomplished in the future work, discussion, or conclusion sections. In this section, we review the future works that the prior works bring up.

\subsection{Overview of Future Works Mentioned}
\citet{hoang-kan-2010-towards} acknowledge their limitation on requiring a pre-existing topic tree as the input. They suggest generating topic trees automatically in their future work. They also suggest automated assistance in interpreting and organizing scholarly work helps build future applications for integration with digital libraries and reference management tools.

\citet{hu2014automatic} note that they will make use of citation sentences to improve their system.

\citet{wang2019toc} consider modeling the generative process of the target paper and the cited papers at a finer granularity. They are particularly interested in taking the citation purpose/intent into consideration. They also mention improving the readability of the generated text.

\citet{deng2021automatic} mention abstractive summarization approaches, entity extraction, and leveraging knowledge bases as their future work.

\citet{abura2020automatic} talk about adding more sentences from multiple papers beyond the title and abstract for consideration. They also mention examining the role of discourse and connecting different citation sentences to a cohesive piece of text.

As the subsequent work of \citet{hu2014automatic}, \citet{xing2020automatic} consider introducing more information like the papers that co-cite the cited papers.

\citet{ge-etal-2021-baco} plan to adapt their multi-input, multi-output framework into more powerful models such as Transformer \cite{vaswani2017attention}. They also hope to extend their citation functions scheme beyond the valence of the citing sentences to more fine-grained categories, such as those outlined in \citet{moravcsik1975some} and 
\citet{lipetz1965improvement}.

\citet{chen-etal-2021-capturing} briefly mentioned that they are interested in the abstract generation and paper generations.

\citet{luu-etal-2021-explaining} mentioned a few future works throughout the paper. Regarding using the citing sentences as partial supervision for their task of explaining the relationship between scientific papers, they suggest filtering or systematically altering in-text citations to be more explanation-like, without changing their approach. They also suggest reusing their SciGPT2 for future work. Moreover, they point out that their language model, SciGen often generates vague and generic sentences, which needs to be improved for automated literature reviews. They further point out that in addition to the problem of missing some relationships, not all citation sentences are useful as explanations. Future work could focus on curating better training sets for their task. Finally, they mention future work should both ensure the factual accuracy of the generated text and improve the modeling of the cited papers.

\subsection{Discussion}
Since the prior works are published throughout the past decade, many future work proposed by the earlier works are already addressed, such as automated topic modeling \cite{hoang-kan-2010-towards} by \citet{hu2014automatic, chen2019automatic, wang2019toc}; citation sentences \cite{hu2014automatic} by the abstractive works \cite{xing2020automatic, ge-etal-2021-baco, luu-etal-2021-explaining}; co-cite papers \cite{xing2020automatic} by \citet{chen2019automatic, wang2019toc}; citation purpose \cite{wang2019toc} by \citet{ge-etal-2021-baco}. There are still many unsolved issues, among which \citet{luu-etal-2021-explaining} mention many of them, such as the generated citation texts are too vague, higher quality of training sets, ensuring factual accuracy and better modeling of the cited papers.

\section{Other Related Studies}

So far, we have reviewed all of the limit number of existing prior studies directly related to the automatic related work generation task. In this section, we zoom out from related work generation to briefly introduce the related tasks that potentially worth integrating to related work generation.

\label{sec:related_studies}
\subsection{Automatic Summarization of Scientific Papers}
Automatic summarization of scientific papers is a super-task that includes the related work summarization, but it has a longer history of studies. \citet{altmami2020automatic} perform an extensive survey of automatic scientific article summarization. They classify scientific article summary into two types: (1) an abstract that provides a general overview of the paper and (2) a summary based on citation sentences. The abstract is not an accurate scientific summary, since it states the contributions in a general and less focused way. It also describes the viewpoint of the author in a biased and incomplete manner \cite{yang2016amplifying}. On the other hand, a citation-based summary leverages a set of citations that co-cite a paper to create a summary of the cited paper \cite{10.5555/1599081.1599168, qazvinian2013generating}. This set of co-citations conveys more information, such as the main findings and more focused contributions of the cited paper than abstract does \cite{elkiss2008blind}. Of course, these co-citations can be biased toward the viewpoints of the citing authors in an incomplete form, as a result, they may not accurately describe the contents of the reference \cite{altmami2020automatic}.

Automatic scientific paper summarization can be divided into two types based on the number of input documents: single document summarization (e.g.  \citet{saggion2000selective, teufel2002summarizing, 10.5555/1599081.1599168, qazvinian2010citation, lloret2013compendium, slamet2018automated}) and multi-document summarization (e.g. \citet{mohammad2009using, khodral2012automatic, erera2019summarization}). Single document summarization generates a summary of a single paper, while multi-document summarization summarizes a set of papers belonging to the same topic jointly in a single summary, which is more challenging than the single document case. The difference between the citation-based multi-document summarization of scientific papers and related work generation is that the former does not necessarily consider the related work section as the generation target. However, because most of the prior approaches of citation-based multi-document summarization of scientific papers are extractive, the actual output is similar to the output of the extractive related work generation systems we reviewed. On the other hand, strictly speaking, there is also no clear borderline between the citation-based multi-document summarization of scientific papers and the abstractive related work generation work that use citation sentences as the generation target, for example, \citet{luu-etal-2021-explaining} phrase their task as ``explaining the relationship between scientific documents'', which does not explicitly work on related work generation.

\subsection{Study of Literature Review Writing}
From information studies' perspective, which is a sub-field of social sciences, \citet{jaidka2010imitating, jaidka2011literature, khoo2011analysis, jaidka2013literature, jaidka2013deconstructing} publish a series of abstracts and experimental papers about their project of literature review writing study. They focus on the analysis of literature review writing itself instead of proposing a system that automatically generates literature reviews. Table \ref{tab:kokil_jaidka} summarizes their novelty and contributions.

\begin{table*}[t]
\begin{center}
    \begin{tabular}{ | l | l | }
    \hline
    \textbf{Prior Work} & \textbf{Novelty \& Contributions} \\ \hline
    \citet{khoo2011analysis} & Classification of integrative and descriptive literature reviews. \\ \hline
    \citet{jaidka2010imitating, jaidka2011literature, jaidka2013literature} & Compare integrative vs. descriptive literature reviews \\ & from multiple aspects.\\ \hline
    \citet{jaidka2013deconstructing} & Propose a template-based system to generate literature reviews. \\ \hline
    \end{tabular}
    \caption{A summary of the study of literature review writing.} \label{tab:kokil_jaidka}
    \vspace{-1em}
\end{center}
\end{table*}

\subsubsection{Macro-scopic Discourse Analysis} \hfill \break
\citet{khoo2011analysis} study 20 papers in information studies to analyze the macro-level discourse structure of literature reviews and identify different styles of literature reviews. They develop a coding scheme to annotate the high-level organization of literature reviews, focusing on the types of information. They use an XML-style annotation scheme to annotate literature reviews with structural element tags, such as \{lit-review, topic, study, description, meta-critique, and meta-summary, etc\}. They also provide an abstract template of the literature review given these discourse tags. Based on this annotation, they largely classify literature reviews into two styles: integrative and descriptive. Descriptive literature reviews summarize individual studies and provide more information on each study, such as methods, results, and interpretation. Integrative literature reviews provide fewer details of individual studies but focus on ideas and results extracted from these papers.

\subsubsection{Integrative v.s. Descriptive Literature Review} \hfill \break

As a concluding paper, \citet{jaidka2013literature} deliver comprehensive results of their studies that cover the work of \citet{jaidka2010imitating, jaidka2011literature}. \citet{jaidka2013literature} manually study 20 papers from the area of information studies, by analyzing the count of the observed cases under different conditions. They compare integrative and descriptive literature reviews mentioned in \citet{khoo2011analysis} from multiple aspects. Intuitively, integrative literature reviews have significantly more research result information and critiques than descriptive literature reviews, while descriptive literature reviews have more research method information introduced. Interestingly, by studying the sections that each type of literature reviews cite, they find that integrative literature reviews reference more information from the results and conclusion sections than descriptive literature reviews. On the other hand, descriptive literature reviews reference more information from the abstract and introduction sections than integrative literature reviews. By comparing the type of information transformation from the cited paper to the literature review sentences used in these two types of literature reviews, including cut-paste, paraphrase, summary, and critical reference, they find that integrative literature reviews have more critique, high-level summarizing, and paraphrasing of source information than descriptive literature reviews. On contrary, descriptive literature reviews have more cut-pasting than integrative literature reviews. This characteristic also holds for the abstract sections that both styles of literature reviews frequently reference. Integrative literature reviews are likely to paraphrase information from the abstract, while descriptive literature reviews are likely to cut-paste information from the abstract. Finally, they find that a large proportion of the research objectives and research methods sentences in either style of the literature review are summarized at a high level.

The observations \citet{jaidka2013literature} reveal valuable characteristics of integrative and descriptive literature reviews for computer scientists who develop automatic related work generation systems to understand the input and output data. The classification of integrative and descriptive styles reflects the rich writing style of literature reviews. However, this classification is still too coarse to further understand literature reviews, since it is a document-level classification instead of paragraph or sentence-level. The sample size of 20 from the information studies field is too small for globally understanding the writing of literature reviews, and the observations may be different if the subject domain of the literature reviews changes.

\subsubsection{Deconstructing Human Literature Reviews} \hfill \break
\citet{jaidka2013deconstructing} apply the analysis results from \citet{khoo2011analysis} and \citet{jaidka2013literature} on 30 information studies papers to propose a template-based system to generate literature reviews. They describe their work at three levels: (1) macro-level document structure; (2) sentence-level rhetorical structure and (3) summarization strategies. 

\citet{jaidka2013deconstructing} first propose a template document structure in the literature review framework. They identify the structures within each topic and their hierarchical relationships in literature reviews by annotating sentences with one or more of the following tags derived from \citet{khoo2011analysis}: \{title, description, meta-summary, meta-critique, current-study, and method \& result\}. These tags suggest rules for generating integrative and descriptive literature reviews introduced in \citet{jaidka2013literature}. Integrative literature reviews should comprise more meta-summary and meta-critique elements, while descriptive literature reviews contain more methods \& results. Structurally, integrative literature reviews are organized as a hierarchical structure with embedded topics. On the other hand, descriptive literature reviews are organized as a flat structure, with many more topic elements per text but fewer embedded topics. An example template integrative literature review has a tree structure. The root node of the tree is a general topic, and its children include a meta-summary node and a few specific topic nodes as siblings. In the third layer, each of the specific topic nodes has children including a meta-summary node and a few study nodes as siblings.

Next, \citet{jaidka2013deconstructing} follow the criteria of \citet{teufel1999argumentative} to identify discourse markers and their templates for functions such as describing a topic or comparing studies. They find 110 expressions in 1298 cases. They apply these findings to develop sentence templates that are associated with the type of literature reviews for text generation by using regular expressions. Finally, they design information selection and summarization strategies by considering the way of information transformation from citing sources to the literature review texts, such as copy-paste or paraphrasing which are introduced in \citet{jaidka2013literature}. They show improvement over MEAD system \cite{radev2004centroid} via ROUGE metric \cite{lin2004rouge} and human evaluation on comprehensibility, readability, and usefulness.

Although \citet{jaidka2013deconstructing} only study 30 papers using a template-based approach, which seems to be insufficient today, their intuition on how literature reviews are generated at the document level is still profound. We believe a strong upgraded version of their system can be developed by replacing their regular expression and manually creating features with state-of-the-art neural network models.

\subsection{Citation Analysis}
Because related work sections contain rich citation texts, analyzing citations naturally becomes a related topic. There has been a line of research on citation analysis, including citation function \cite{garfield1965can, teufel2006automatic, dong2011ensemble, jurgens2018measuring, tuarob2019automatic, zhao2019context}, citation intent \cite{cohan2019structural, lauscher2021multicite}, and citation sentiment \cite{athar2011sentiment, athar-teufel-2012-context, ravi2018article} etc. Table \ref{tab:citation_analysis} summarizes the datasets proposed by the surveyed works.

\begin{table*}[t]
\begin{center}
    \begin{tabular}{  l  l  l l }
    \hline
    \textbf{Prior Work} & \textbf{Source Domain} & \textbf{Label Set} & \textbf{Size} \\ \hline
    Citation Function & & & \\ \hline
    \citet{teufel2006automatic} & Computational & \{Weakness, Agreement/usage/ & 548 \\ 
    & linguistics (ACL, EACL, & compatibility (6 categories),  & \\
    & ANLP, COLING) & Contrast (4 categories), Neutral\} & \\ \hline
    \citet{dong2011ensemble} & ACL Anthology & \{Background, Fundamental idea, & ? \\
    & & Technical basis, Comparison\} & \\ \hline
    \citet{jurgens2018measuring} & ACL Reference Corpus & \{Background, Uses, Compares & 1969 \\
    & (ARC) & or Contrasts, Motivation, & \\ 
    & & Continuation, Future\} & \\\hline
    \citet{tuarob2019automatic} & AI, algorithm and & \{Use, Extend, Mention, & 8796 \\ 
    & theory, data mining, & NotAlgo\} & \\
    & parallel computing, & & \\
    & security, HCI, & & \\ 
    & bioinformatics & & \\\hline
    \citet{zhao2019context} & ARC, NeurlPS, PubMed & \{Material (Data); Method (Tool, & 3088 \\
    & & Code, Algorithm); Supplement & \\
    & & (Document, Website, Paper, & \\
    & &  License, Media)\} & \\
    & & \{Use, Produce, Introduce,  & \\
    & & Extend, Other, Compare\} & \\ \hline
    Citation Intent & & & \\ \hline
    \citet{cohan2019structural} & Computer Science, & \{Background, Method, & 11020 \\
    & Medicine & ResultComparision\} & \\ \hline
    \citet{lauscher2021multicite} & ACL Anthology & \{Background, Motivation, Uses,  & 12653 \\
    & & Extends, Similarities, Differences,  & \\
    & & Future Work\} & \\ \hline
    Citation Sentiment & & & \\ \hline
    \citet{athar2011sentiment} & ACL Anthology & \{positive, negative, objective\} & 8736 \\ \hline
    \citet{athar-teufel-2012-context} & ACL Anthology & \{positive, negative, objective\} & 1741 \\
    \hline
    \end{tabular}
    \caption{A summary of datasets from prior works about citation analysis.} \label{tab:citation_analysis}
    \vspace{-1em}
\end{center}
\end{table*}

\subsubsection{Citation Function and Intent} \hfill \break
\citet{teufel2006automatic} is the first-ever work on annotation of citation function in scientific texts. They define citation function as the author's reason for citing a given paper. They propose an annotation scheme for citation functions with the labels of (1) explicit statement of weakness; (2) contrast or comparison with other work (4 categories); (3) agreement/usage/compatibility with other work (6 categories); and (4) a neutral category. Their data are obtained from 360 conference articles in the computational linguistics domain, including annotated 548 citations. They take features such as cue phrases and use the IBK algorithm (implemented by \citet{witten2005practical}), which is an implementation of k-nearest-neighbor. Because their proposed label set is unbalanced, they also merge labels to form a coarse label set with only \{Weakness, Positive, Contrast, Neutral\} labels, which results in higher macro-F1 performance.

\citet{dong2011ensemble} annotate citations in the papers from ACL Anthology with citation functions including \{Background, Fundamental idea, Technical basis, Comparison\}. Their created features including textual features based on cue words, physical features such as location or popularity of the citation, and POS-based syntactic features. They use an ensemble of algorithms such as BayesNet and NaiveBayes and applied self-training mechanism \cite{zhu2009introduction}, which includes high-confidence automatically annotated data points into the training data.

\citet{jurgens2018measuring} annotate citation functions on a corpus of 1436 citations from 52 papers and 533 supplemental contexts from 133 papers drawn from the ACL Anthology Reference Corpus (ARC) \cite{bird2008acl}. Their labels are more concise, including \{Background, Uses, Compares or Contrasts, Motivation, Continuation, Future\} in decreasing order of frequency. To overcome the class imbalance, they up-sample less frequent classes. Their best model is implemented with a feature-rich Random Forest classifier. \citet{jurgens2018measuring} further apply their citation function labels to perform a series of analyses including the narrative structure of citation function, venues and citation patterns, the evolution of venues, predicting future impact, and the growth of rapid discovery science.

\citet{tuarob2019automatic} zoom their attention to the citation function for algorithms. They annotate 8796 citations from multiple computer science disciplines such as AI, algorithm and theory, data mining, parallel computing, security, HCI, and bioinformatics. They propose algorithm domain-specific citation function labels: \{Use, Extend, Mention, NotAlgo\}. They create various features including structures, lexical, morphological and grammatical, cue words, sentiment, and venue. They also consider content features such as TF-IDF and used data balancing techniques to overcome the class imbalance. They use an ensemble of multiple base classifiers such as Random Forest, Convolutional Neural Networks \cite{lecun2015deep} to train on the 4-label classification task, as well as binary classification tasks for each of the labels.

\citet{zhao2019context} propose a new task of modeling the role and function for online resource citations in scientific literature. They construct a dataset called SciRes, which includes 3088 manually annotated examples based on paper full texts from ARC, NeurlPS, and PubMed. Their examples are called resources contexts, which are word sequences surrounding the resource citation. They annotate each resource citation with fine-grained and coarse category role types, as well as resource function types. The category roles include \{Material (Data); Method (Tool, Code, Algorithm); Supplement (Document, Website, Paper, License, Media)\}, and the resource function types include \{Use, Produce, Introduce, Extend, Other, Compare\}. They overcome the class imbalance by up-sampling minority class samples when training models. They jointly fine-tune BERT \cite{devlin2018bert} on the coarse and fine-grained category role labels as well as the resource function labels with multi-task learning for the classification task (SciResCLF). They further apply the architecture of SciResCLF as a recommendation system (SciResREC) for a resource ranking problem.

\citet{cohan2019structural} work on the problem of identifying the intent of a citation in scientific papers. They create a SciCite dataset by annotating 11020 citation sentences from 6627 papers in computer science and medicine extracted from the Semantic Scholar corpus \footnote{\url{https://semanticscholar.org/}}. They only have a simple annotation scheme consisting of three labels: \{Background, Method, ResultComparision\}. They use a bidirectional LSTM \cite{hochreiter1997long} (BiLSTM) with GloVe \cite{pennington2014glove} and ELMo \cite{peters2018deep} as word embeddings. They leverage citation section title and citation worthiness, which refers to whether a given sentence is a citation or not, as auxiliary scaffolding tasks for multi-task learning. They also show some improvements using their method on the ACL-ARC citations dataset proposed by \citet{jurgens2018measuring}.

\citet{lauscher2021multicite} question the veracity of the traditional citation analysis framework, where each sentence is only assigned with a single intent, and each intent is only assigned to one sentence (single-sentence single-intent). Instead, they propose a multi-sentence multi-intent framework, where each sentence may be assigned with multiple intent labels, and each intent may correspond to multiple sentences. Under this new framework, they propose a new dataset called MultiCite from the computational linguistics sub-partition of S2ORC \cite{lo-wang-2020-s2orc} dataset. Their labels contain \{Background, Motivation, Uses, Extends, Similarities, Differences, Future Work\} derived from \citet{jurgens2018measuring}. They try both SciBERT \cite{beltagy2019scibert} and RoBERTa \cite{liu2019roberta}, on top of which they place a multi-label classification head consisting of a set of sigmoid classifiers (one for each of the classes) to perform the classification task. Furthermore, they formulate citation analysis as a question answering (QA) task, by asking a binary answer given a citation paper pair and a candidate citation intent label. They use a multi-task sequence-to-sequence Longformer-Encoder-Decoder (LED) \cite{Beltagy2020Longformer} model for this QA task.

\subsubsection{Citation Sentiment} \hfill \break
\citet{athar2011sentiment} manually annotate 8736 citations from 310 research papers taken from the ACL Anthology \cite{bird2008acl}. They identify citation references with regular expressions and label each sentence as \{positive, negative, objective\}. They use feature-based Support Vector machine (SVM) \cite{cortes1995support} for the classification task.

Later, \citet{athar-teufel-2012-context} extend \citet{athar2011sentiment}'s work with context sentences. They annotate 1741 citations from a subset of \citet{athar2011sentiment}'s dataset by considering two sentences each before and after an explicit citation reference. By considering the context, they identify significantly more citation sentiments labels than \citet{athar2011sentiment}, mainly because more implicit citations (e.g. ``this method'', ``the method'') are labeled.


\subsection{Discourse Analysis} 
Zooming out from citation sentences to general scientific sentences, there has been a significant amount of work on understanding the rhetorical components of scientific discourse \cite{teufel1999discourse, teufel2002summarizing, hirohata2008identifying, liakata-etal-2010-corpora, liakata2012automatic, de2012verb, burns2016automated, PubMed-RCT, huang2020coda, li2021scientific}. Table \ref{tab:discourse_analysis} summarizes the discourse analysis studies surveyed.

\begin{table*}[t]
\begin{center}
    \begin{tabular}{  l  l  l l }
    \hline
    \textbf{Prior Work} & \textbf{Source Domain} & \textbf{Label Set} & \textbf{Size} \\ \hline
    \citet{teufel1999argumentative}, & Computational & \{Background, Other, Own, Aim,   & 1248 \\ 
    \citet{teufel1999discourse, teufel2002summarizing} & linguistics & Textual, Contrast, Basis\}  & sentences \\\hline
    \citet{hirohata2008identifying} & Medline & \{objective, methods, & 51000 \\
    & (Automated) & results, conclusions\} & abstracts\\\hline
    \citet{PubMed-RCT} & PubMed & \{Objective, Background, Methods,  & 331k \\
    & (Automated) & Results, Conclusions\} & abstracts\\ \hline
    \citet{liakata-etal-2010-corpora} & Physical chemistry, & \{Motivation, Goal, Object, & 265 \\
    (ScoreSC) & biochemistry & Method, Experiment, Observation,  & papers \\
    & & Result, Conclusion\} & \\ \hline
    \citet{liakata-etal-2010-corpora} & Royal Society & \{Aim, Nov\_Adv, Co\_Gro & 61 \\
    (AZ-II) & of chemistry & Othr, Prev\_Own, Own\_Mthd,  & papers \\
    & & Own\_Fail, Own\_Res, Own\_Conc, & \\
    & & CoDi, Gap\_Weak, Antisupp, & \\
    & & Support, Use, Fut \} & \\ \hline
    \citet{de2012verb} & Biology & \{Motivation, Goal, Object, & 399 \\
    & & Method, Experiment, Observation, & clauses \\
    & &  Result, Conclusion\} & \\ \hline
    \citet{burns2016automated} & INTACT & \{Motivation, Goal, Object, & 76 \\
    & (biology) & Method, Experiment, Observation, & articles \\
    & &  Result, Conclusion\} & \\ \hline
    \citet{li2021scientific} & INTACT & \{Motivation, Goal, Object, & 6124 \\
    & (biology) & Method, Experiment, Observation, & clauses \\
    & &  Result, Conclusion\} & \\
    \hline
    \end{tabular}
    \caption{A summary of datasets from prior works about discourse analysis.} \label{tab:discourse_analysis}
    \vspace{-1em}
\end{center}
\end{table*}

\citet{teufel1999argumentative, teufel1999discourse, teufel2002summarizing} propose a pioneering scheme called ``Argumentative Zoning'' (AZ) to segment scientific articles into zones of discourse-level argumentation. They manually annotate an AZ corpus in which each zone (block of texts) is labeled with one of the labels from \{Background, Other, Own, Aim, Textual, Contrast, Basis\}. They further use a naive Bayesian classifier to perform the classification task of AZ. As the application of AZ, they propose to apply AZ for document summarization and citation analysis.

\citet{hirohata2008identifying} categorize sentences in Medline abstracts \footnote{\url{https://www.nlm.nih.gov/bsd/medline.html}} into four sections: \{objective, methods, results, conclusions\}. They formulate the categorization task as a sequence labeling problem and employ conditional random fields (CRFs) \cite{lafferty2001conditional} to perform the task. Their training corpus is acquired automatically from Medline abstracts, where the authors state the section labels explicitly. 

Similarly, \citet{PubMed-RCT} collect 331k of abstracts from PubMed that are explicitly labeled with section heads (\{Objective, Background, Conclusions, Methods, Results\}) by the authors.

\citet{liakata-etal-2010-corpora} propose two complementary annotation schemes for sentence-based annotation of full scientific papers, CoreSC and AZ-II. Both of them are applied to research articles in chemistry. CoreSC annotation scheme consists of three layers. The first layer includes concepts from CISP \cite{soldatova2007ontology}: \{Motivation, Goal, Object, Method, Experiment, Observation, Result, Conclusion\}. The second layer is for the annotation of properties of the concepts (e.g. ``New'', ``Old''). The third layer caters to concept ID. AZ-II extends AZ \cite{teufel1999argumentative, teufel1999discourse, teufel2002summarizing} to sentence level with more fine-grained 15 labels. They not only annotate CoreSC corpus and AZ-II corpus but also compare the label distributions of a corpus jointly annotated with both CoreSC and AZ-II schemes. Later, \citet{liakata2012automatic} work on the automatic recognition of the CoreSC scheme, using SVM \cite{cortes1995support} and CRF \cite{lafferty2001conditional}.

\citet{de2012verb} propose a discourse segment type scheme for biology experimental papers, and study the correlation between the discourse segment type and the tense used from a linguistic perspective. They define the labeling rule of tagging \{Goal, Fact, Result, Hypothesis, Method, Problem, Implication\} to each clause in biology papers.

\citet{burns2016automated} annotate 76 biology articles collected from INTACT database \cite{orchard2013mintact} with three types from labels: (1) Discourse segment types, a.k.a. SciDT \cite{dasigi2017experiment, li2021scientific}, defined by \citet{de2012verb}; (2) The description in the texts linked to each experimental sub-figure, which is later named as evidence fragment \cite{burns2017extracting}; and (3) The type of biological experiments in each paragraph. They use feature-based SVM \cite{cortes1995support} and CRF \cite{lafferty2001conditional} to perform the task of discourse segment type classification.

The tasks derived from \citet{burns2016automated}'s dataset are further addressed by subsequent works. \citet{dasigi2017experiment} formulate the discourse tagging task as a sequence labeling task and improve the performance using an attention mechanism containing an Recurrent Neural Network (RNN) followed by an LSTM \cite{hochreiter1997long}. \citet{burns2017extracting} propose a rule-based approach to detect the evidence fragment. \citet{li2021scientific} further improve these tasks by performing transfer learning from scientific discourse tagging tasks, which cover SciDT \cite{burns2016automated, dasigi2017experiment, li2021scientific}, PubMed-RCT \cite{PubMed-RCT} and CODA-19 \cite{huang2020coda}, to down-stream tasks such as evidence fragment detection \cite{burns2016automated, burns2017extracting}. The experimental type classification is improved by \citet{burns2019building} with LSTM \cite{hochreiter1997long} and CNN \cite{lecun2015deep}, and further improved by \citet{wu2020melinda} with multi-modal Transformers \cite{su2019vl, lu2019vilbert}. 

\subsection{Cited Text Span}

\subsubsection{Explicit Annotation of Cited Text Span}
 \citet{aburaed-etal-2020-multi} extend \citet{hoang2010towards}'s RWSData dataset by annotating the Cited Text Spans (CTS) \cite{wang2019toc}. They annotate the specific sentences in reference papers that each citation in the target paper cites. For each of these reference papers, they further collect a set of papers that also cite these papers. They also implement several automatic systems that retrieve the CTS given the citation context.
 
Unfortunately, because the CTS is not recorded by the current rule of citation network, and due to the natural ambiguity of the CTS, \citet{aburaed-etal-2020-multi}'s inter-annotator agreement of the CTS annotation is low, which only has the average pairwise agreement of 0.52. As a result, they release a refined version of the dataset that removes all controversial annotations. Consequently, the precision of their CTS retrieval systems is also unsatisfactory. Nonetheless, their effort on the explicit annotation of CTS is profound for automatic related work generation. 

\subsubsection{Shared Tasks of Cited Text Span Identification}
\citet{jaidka2018insights, jaidka2019cl} propose the CL-Scisumm shared task in the domain of computational linguistics papers. They expand their dataset year by year, which has the task of 1a): For each citing sentence, identify CTS in reference papers that are of the granularity of a sentence fragment, a full sentence, or no more than 5 of consecutive sentences. 1b): For each CTS, identify the type of facet from a predefined set. 2): Generate a summary from the CTS. \citet{yasunaga2019scisummnet} further expand this dataset to ScisummNet corpus, which consists of 1000 papers. They also advocate the importance of considering both the CTS and the cited paper's abstract to integrate the viewpoints from multiple sources. This CL-Scisumm task \cite{jaidka2018insights, jaidka2019cl, yasunaga2019scisummnet} provides a valuable pinpointed information source for automatic related work generation.

\section{Discussions}
\label{sec:discussion}
\subsection{Extractive Related Work Generation}
The earliest work by \citet{hoang2010towards} is published in 2010, before the era that neural networks prevail. Due to the technical limitation of document summarization techniques, early solutions for related work generation are all extractive approaches. Sentences as the basic unit from the rest of the target paper and the cited papers are extracted with the help of a manually or automatically generated topic model to construct the entire related work section. Because the sentences are not generated by the systems, the designs of these systems mainly focus on how to extract sentences to form coherent, topic-biased, and comprehensive related work texts. 

Extractive related work generation systems usually produce natural (human-generated) individual sentences with high ROUGE scores. However, as \citet{ge-etal-2021-baco} points out, related work sections are not a simple aggregated summary of each cited paper. Due to the naive aggregation of the extracted sentences, the output related work sections lack coherence and have low fluency in human evaluated scores, as we compare in \cref{sec:evaluations}. As \citet{ge-etal-2021-baco} further criticize, these extractive approaches may introduce intellectual property issues, since sentences from prior works are copied to the generated related work section without modifications.

\subsection{Abstractive Related Work Generation}
Neural network-based abstractive related work generation only emerges recently as the neural network models such as pointer-generator network \cite{see-etal-2017-get} and Transformer \cite{vaswani2017attention} demonstrate their success on abstractive summarization in general domains such as the news domain. All of the works introduced above show that the attention mechanism is effective in fusing the representations of multiple documents via the Pointer-Generator Networks \cite{see-etal-2017-get} and Transformers \cite{vaswani2017attention}. We also believe that attention-based neural network models will become a prevalent approach for related work generation in future studies.

Additionally, the studies introduced above provide some take-home messages that apply to future explorations of related work generation: (1) Combine manual and automatic approaches for efficient data collection. Due to the high cost of manual annotation, training a model that can perform the annotation task well enough given a small set of the manually annotated dataset and predicting this model on unlabeled inputs can quickly produce a large-scale dataset for training neural sequence generation models. (2) External supervision signals such as citation network, citation function, and salience estimation improve the citation sentence generation performance \cite{ge-etal-2021-baco}. (3) More sophisticated multi-document encoder such as RRG \cite{chen-etal-2021-capturing} helps. (4) Proper input selection helps improve the performance \cite{luu-etal-2021-explaining}.

However, the works introduced above are merely the early exploration of related work generation. The state-of-the-art system is still far from satisfactory. Currently, even as the first step, the definition of related work generation task is not widely recognized yet. Each work defines their task, collects their dataset, and barely (except \citet{luu-etal-2021-explaining}) release their code so that it is challenging to perform a parallel comparison across prior works. Furthermore, all of the existing works attempt to solve related work generation as a variation of multi-document summarization tasks, but this may not be the optimal solution due to the complex nature of the related work section, which requires deep understanding and thorough digestion of prior works.

\subsection{General Discussions}
There is no doubt that automatic related work generation is still at the early stage of exploration, as the proposed system performance is still far from satisfactory for users to use them to assist the composition of related work sections in academic writing. More importantly, we uncover several issues about the study of automatic related work generation within this meta-study.

Automatic related work generation has been regarded as the intersection of scientific document summarization as multi-document summarization. Naturally, all of the related work generation studies reviewed aim to propose new systems by following similar patterns: obtaining a dataset; developing extractive approaches that select salient sentences from inputs and simply concatenate them as the output, or developing abstractive approaches that generate citation sentences or paragraphs, and finally conducting automatic and manual evaluations. Although this train of thought is a valid way to approach the related work generation problem, it is not the only valid one. There are still many unsolved issues that we will briefly discuss below, improving or solving any of which may result in a valuable study. 

As we previously pointed out, almost all studies reviewed collect their dataset from scratch, without reusing the datasets published in previous studies, due to various reasons, presumably different task definitions, or low quality of the released datasets. This issue is probably because all these prior works mainly focus on improving the system performance itself, and pay less attention to collecting high-quality datasets that are reusable by the subsequent studies. As a result, most of the reviewed studies are barely comparable with each other, because all of the task definitions, datasets, and performance evaluation metrics are more or less different. For example, as the latest extractive study, \citet{deng2021automatic} only compare against the MEAD \cite{radev2004centroid} baseline, which is not persuasive how well their system performs. It is urgent to create a standardized shared task such as CL-Scisumm \cite{jaidka2018insights, jaidka2019cl} to avoid the repetitive data collection and ineffective performance evaluation process.

Another serious outcome of the fixed train of thought for related work generation studies is most studies simply aim to improve the scores measured by their chosen evaluation metrics under certain assumptions without paying much attention to the practicality of their systems. For example, almost all of the prior works reviewed use sentences as their basic unit: they either select salient sentences to form a summary as the related work section, or aim to generate citation sentences each pass. Despite the simplicity of sentences to extract and process, they may not be the optimum basic unit, since practically one citation sentence may include multiple citations from multiple cited papers, or multiple consecutive sentences may describe one single cited work. Another example of the assumption is most reviewed studies use abstracts or sometimes conclusion sections as the representations of the entire cited papers, which is another over-simplification of the problem. Although the authors may defend by claiming that the current computational resources cannot support the processing of the full-text papers, it is still a necessary direction to develop efficient encoding methods to represent the full-text of the involved papers involved.

Furthermore, there are more unsolved critical issues as the reviewed studies mentioned in their future work. As \citet{luu-etal-2021-explaining} point out, it is critical to improve the factuality of the generated citation texts, since even though the Transformer-based \cite{vaswani2017attention} models generate fluent texts, they also generate hallucinated phrases that are inconsistent with the citation contexts and the cited papers. Even worse, due to the cognitive challenge of human evaluation of the generated citation texts, the factuality aspects are much harder to judge than other aspects such as fluency. This factuality issue may be mitigated by leveraging the knowledge base as suggested by \citet{deng2021automatic}, but the available resources and practical methods of leveraging the knowledge base for summarization tasks are very limited. Moreover, existing studies are mostly satisfied with obtaining a collection of properly phrased citation sentences. This is still halfway to the full version of the readable related work section. The study of how to connect sentences coherently to a long text, and simultaneously achieve the purpose of generating a factual, informative and succinct related work section is missing entirely. Finally, due to the intellectual challenge for understanding and processing scientific documents, the majority of texts studied for related work generation are focused on the computational linguistics that the authors are naturally familiar with. It is a challenging but necessary task to expand the domain of texts to other various domains such as natural sciences and social sciences.

\section{Conclusion}
We present a critical review of the state-of-the-art studies about the task of automatic related work generation by conducting a meta-study that compares across the limited number of available prior works from the aspects of problem formulation, dataset collection, approaches, result and evaluation, and future work. We aim to help readers with sufficient basic knowledge about NLP but limited experience on automatic related work generation quickly grasp the background, identify and tackle the limitations of the previous studies, and pinpoint the potential novelty of their studies. Our study find that the following approaches effectively improves the related work generation system: (1) Combining manual and automatic approach for efficient data collection; (2) Adding external supervision signals such as citation network, citation function, and salience estimation; (3) Developing more sophisticated multi-document encoders; (4) Filtering proper inputs. We also identify several issues and to-dos to be solved in future work: (1) Lacking focused studies on a specific aspect of related work generation; (2) Lacking a standardized task definition and dataset; (3) Exploring solutions that do not rely on the assumptions that take sentences as the basic unit, or only using part of the paper as the representation of the full paper; (4) The issue of losing factuality, and hallucination of the generated texts; (5) Lacking knowledge-driven approaches; (6) Lacking approaches to construct a full related work section; (7) Extending text-domain from computational linguistics to other domains.


\bibliographystyle{ACM-Reference-Format}
\bibliography{sample-base}

\appendix

\end{document}